\RequirePackage{fix-cm}
\documentclass[twocolumn,natbib]{svjour3}          % twocolumn
\smartqed  % flush right qed marks, e.g. at end of proof
\usepackage{graphicx}
\usepackage{newtxtext,newtxmath} % use Times fonts if available on your TeX system
\usepackage{subfigure,multirow,array,amsmath,booktabs}

\usepackage{marvosym}
\usepackage[colorlinks,citecolor=blue]{hyperref}

\usepackage[flushleft]{threeparttable}
\newcommand{\highlight}[1]{\textcolor{black}{#1}}

\journalname{International Journal of Computer Vision}
\begin{document}
\begin{sloppypar}

% \title{Facial Action Unit Detection via Self-Attention Constraining and Sample Deconfounding
% }
\title{Facial Action Unit Detection by Adaptively Constraining Self-Attention and Causally Deconfounding Sample
}
% \subtitle{Do you have a subtitle?\\ If so, write it here}

%\titlerunning{Short form of title}        % if too long for running head

\author{Zhiwen~Shao$^{1,2,3,4}$ \and
        Hancheng~Zhu$^{1,3}$ \and
        Yong~Zhou$^{1,3}$ \and
        Xiang~Xiang$^2$ \and
        Bing~Liu$^{1,3}$ \and
        Rui~Yao$^{1,3}$ \and
        Lizhuang~Ma$^4$
}

%\authorrunning{Short form of author list} % if too long for running head

\institute{
          Hancheng~Zhu (\Letter) \at
          \email{zhuhancheng@cumt.edu.cn} \and
          Yong~Zhou (\Letter) \at
          \email{yzhou@cumt.edu.cn} \and
          Xiang~Xiang (\Letter) \at
          \email{xxiang@cs.jhu.edu
          }
%         \emph{Present address:} of F. Author  %  if needed
          \and
          $^1$School of Computer Science and Technology, China University of Mining and Technology, Xuzhou 221116, China\\
          $^2$Key Laboratory of Image Processing and Intelligent Control (Huazhong University of Science and Technology), Ministry of Education, Wuhan 430074, China\\
          $^3$Mine Digitization Engineering Research Center of the Ministry of Education, Xuzhou 221116, China\\
          $^4$Department of Computer Science and Engineering, Shanghai Jiao Tong University, Shanghai 200240, China
}

\date{Received: date / Accepted: date}
% The correct dates will be entered by the editor

\maketitle

\begin{abstract}
Facial action unit (AU) detection remains a challenging task, due to the subtlety, dynamics, and diversity of AUs. Recently, the prevailing techniques of self-attention and causal inference have been introduced to AU detection. However, most existing methods directly learn self-attention guided by AU detection, or employ common patterns for all AUs during causal intervention. The former often captures irrelevant information \highlight{in a global range}, and the latter ignores the specific \highlight{causal} characteristic of each AU. In this paper, we propose a novel AU detection framework called AC$^{2}$D by adaptively constraining self-attention \highlight{weight} distribution and causally deconfounding the sample confounder. Specifically, we explore the mechanism of self-attention \highlight{weight} distribution, in which the self-attention \highlight{weight} distribution of each AU is regarded as spatial distribution and is adaptively learned under the constraint of location-predefined attention and the guidance of AU detection. Moreover, we propose a causal intervention module for each AU, in which the bias caused by training samples and the interference from irrelevant AUs are both suppressed. Extensive experiments show that our method %significantly
% outperforms
\highlight{achieves competitive performance compared to} state-of-the-art AU detection approaches on challenging benchmarks, including BP4D, DISFA, GFT, and \highlight{BP4D+} in constrained scenarios and Aff-Wild2 in unconstrained scenarios. The code is available at https://github.com/ZhiwenShao/AC2D.

% Adaptive constraining \and Self-attention distribution
% \PACS{PACS code1 \and PACS code2 \and more}
% \subclass{MSC code1 \and MSC code2 \and more}
\end{abstract}

% \begin{sloppypar}

\section{Introduction}
In recent years, facial action unit (AU) detection has gained significant attention in the fields of computer vision and affective computing~\citep{li2018eac,niu2019local,shao2021explicit}. AU detection involves the recognition of subtle facial movements that correspond to specific emotional expressions. Each AU is linked to one or more local muscle actions, as defined by the facial action coding system (FACS)~\citep{ekman1978facial,ekman2002facial}. With the aid of deep learning technology, the performance of AU detection has been significantly improved~\citep{jacob2021facial,chen2022causal,shao2023facial}. However, AU detection remains a challenging task since some inherent characteristics are not thoroughly exploited.

Inspired by the power of prevailing transformer~\citep{vaswani_attention_2017}, some works introduce the self-attention mechanism to AU detection. For instance, \cite{jacob2021facial} and \cite{wang2022action} adopted a convolutional network to extract the feature of each AU, then input AU features to a transformer for correlational modeling among AUs. In these methods, self-attention is used as an application and is learned only under the guidance of AU detection, in which the characteristics including subtlety, dynamics, and diversity of AUs are difficult to be modeled. %, due to the limited dataset diversity and the .
Besides, \highlight{convolutional attention weight distribution has been explored in AU detection and has significantly enhanced the performance~\citep{li2018eac,jacob2021facial,shao2019facial,shao2023facial}. However,} the research about the inherent mechanism of \highlight{transformer attention (also known as self-attention) weight} distribution is ignored.
%researches about the mechanism regards to the distribution of self-attention are ignored.

\begin{figure}
\centering\includegraphics[width=0.91\linewidth]{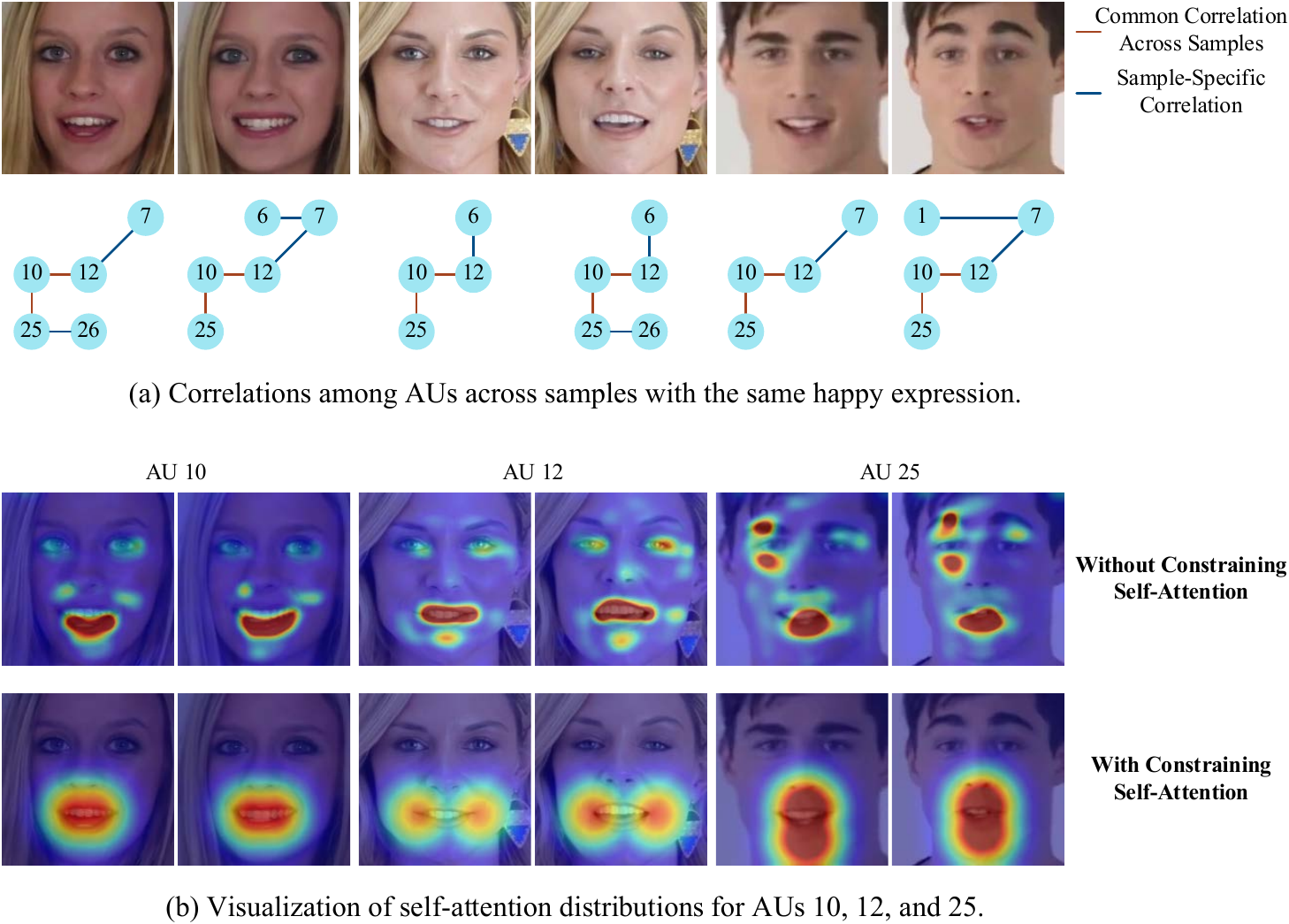}
\caption{Illustration of AU correlations and self-attention \highlight{weight} distribution on sample images from Aff-Wild2~\citep{kollias2019expression,kollias2021analysing} with the same happy expression. In (a), AU co-occurrences contain common co-occurrence of AU 10 (upper lip raiser), AU 12 (lip corner puller), and AU 25 (lips part) across samples, as well as sample-specific AU co-occurrences. In (b), we visualize the average self-attention \highlight{weight} distribution of example AUs 10, 12, and 25 for our method without constraining self-attention and with constraining self-attention. The self-attention \highlight{weight} distribution is visualized as spatial distribution, in which attention weights are overlaid on the sample image for better viewing.}
%are drawn using colors in the color bar and
\label{fig:motivation}
\end{figure}

Since the appearances of AUs and the correlations among AUs are sometimes different across samples, most AU detection methods suffer from \highlight{predicting} bias, %for unseen samples,
\highlight{in which the prediction of AU occurrences/non-occurrences biases to frequently seen or easily modeled samples.}
% in which the predictions of AU occurrences/non-occurrences bias to frequently seen or easily modeled AU occurrences/non-occurrences
Recently, \cite{chen2022causal} employed causal inference theory~\citep{pearl2000models,rubin2005causal} to remove the bias caused by variations across subjects, which is a pioneering work of causal inference based AU detection.
% has gained attention in the field of AU detection.
This method requires identity annotations of training data, and employs a common causal intervention module for all AUs. However, %a same
\highlight{the} same
subject still often presents %a same
\highlight{the} same
AU with different appearances and correlations in different time or scenarios, and each AU has specific \highlight{causal} characteristic. For example, six samples with happy expression in Fig.~\ref{fig:motivation}(a) all show different co-occurrences of AUs, in which each \highlight{pair of two adjacent} samples \highlight{belongs} to the same subject. \highlight{Therefore, it is desired to model the causalities in more fine-grained sample level.}

To tackle the above \highlight{issues}, %limitations,
we propose an end-to-end AU detection framework named \textbf{AC$^{2}$D} by exploring the mechanism of self-attention \highlight{weight} distribution and removing the \highlight{predicting} bias from variations across samples. In particular, we simplify the structure of ResTv2~\citep{zhang2022rest} to be the backbone of our framework, in which two stages are used to extract rich feature shared by AUs, and then each AU uses one stage as its specific branch. In each AU branch, we reshape the scaled dot-product attention~\citep{vaswani_attention_2017} to spatial attentions with multiple channels, and encourage the average spatial attention over channels to close to an attention map predefined by AU locations.

As shown in Fig.~\ref{fig:motivation}(b), the learned self-attention of a certain AU without constraining already have some high responses near the AU region, which demonstrates explaining self-attention from the perspective of spatial attention is reasonable. Since the learning of scaled dot-product attention is also guided by AU detection during training, the self-attention \highlight{weight} distribution is adaptively constrained, in which both accurate feature learning from prior knowledge about AU locations and strong modeling ability from automatic self-attention learning are exploited. \highlight{In this way, the constrained self-attention can capture AU related local information while preserving global relational modeling capacity.}
% the  about the location of each AU
%by facial landmarks

Moreover, we propose to remove the negative impacts from sample confounder with inherent sample characteristics. %and specific image scenes.
As illustrated \highlight{by different samples with the same expression} in  Fig.~\ref{fig:motivation}(a), the co-occurrence of AUs 10, 12, and 25 is common across samples, while other %AU occurrences are determined by individual samples. Such %such as image illuminations, image backgrounds
\highlight{sample-specific AU occurrences are determined by sample characteristics including the time and scenario recording the sample and subject custom of exhibiting the expression}. Specifically, we use a causal diagram to formulate the causalities among facial image, sample \highlight{confounder}, and AU-specific occurrence probability. Then, we design a causal intervention module to deconfound the sample confounder for each AU by introducing backdoor adjustment~\citep{pearl2016causal}. In our framework, adaptive constraining on self-attention \highlight{weight} distribution and causal deconfounding of sample confounder are jointly trained.
% in which these three AUs are all located around the mouth region. In Fig.~\ref{fig:motivation}(b), after constraining the self-attention, the average self-attention distributions of the three AUs are all only highlighted around the mouth region. This demonstrates that only common correlations across samples are emphasized, which is consistent with the goal of sample deconfounding.

The main contributions of this work are threefold:
\begin{itemize}
    \item We investigate self-attention \highlight{weight} distribution from the perspective of spatial attention, and propose to adaptively constrain self-attention, \highlight{in which local subtle information associated with each AU is captured while global relational modeling ability is preserved}.
    % which integrates the advantages of both prior knowledge and self-attention learning.
    To our knowledge, this is the first work of exploring the mechanism of self-attention \highlight{weight} distribution in the AU detection field.

    \item We formulate the causalities among image, sample \highlight{confounder}, and AU-specific occurrence probability via a causal diagram, and propose to deconfound the sample confounder in the prediction of each AU by causal intervention. \highlight{This is beneficial for reasoning AU-specific causal effects and suppressing the predicting bias caused by sample variations.}
    \item Extensive experiments on benchmark datasets demonstrate that our approach
    % outperforms the state-of-the-art AU detection works
    \highlight{achieves comparable performance} in terms of both constrained scenarios and unconstrained scenarios.%significantly
\end{itemize}
% \vspace{-0.25in}

\begin{figure*}
\centering\includegraphics[width=\linewidth]{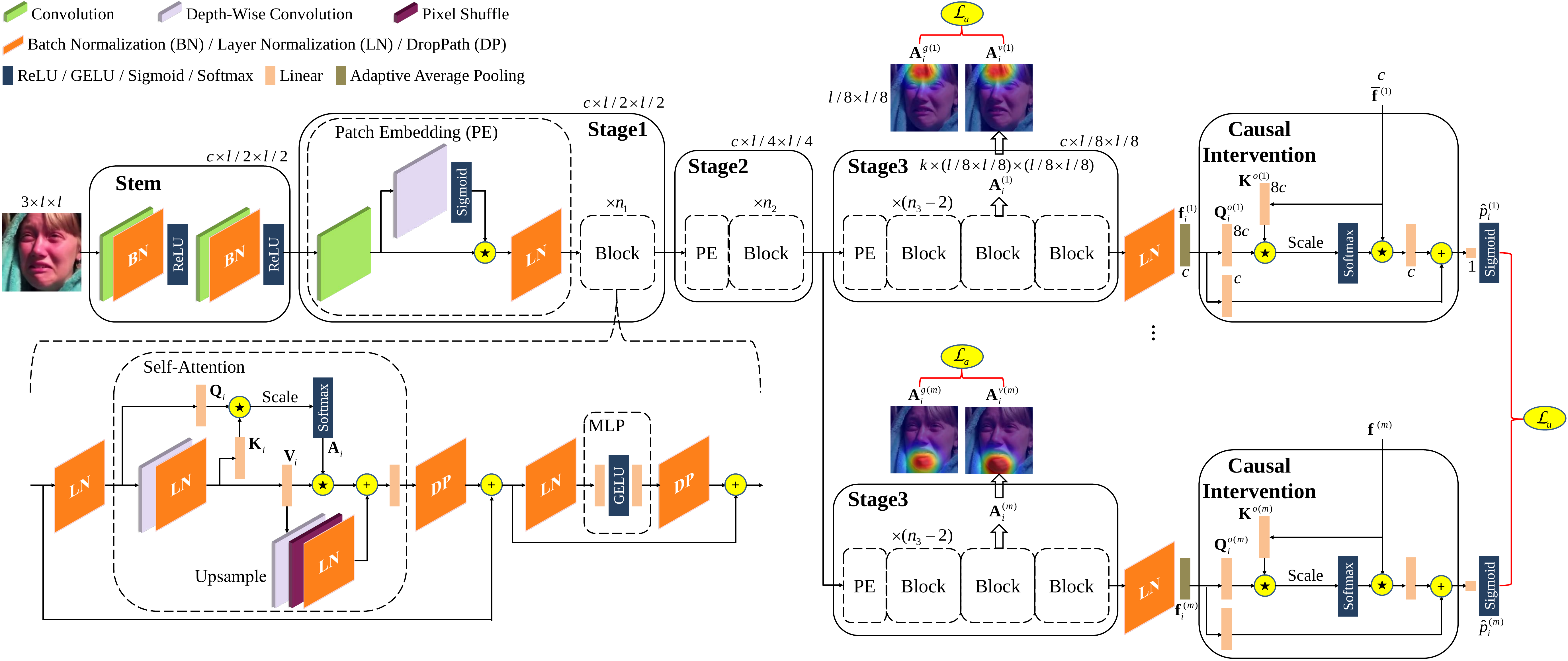}
\caption{The architecture of our AC$^{2}$D framework, which uses a simplified structure of ResTv2~\citep{zhang2022rest}. Given the $i$-th sample image in the training set, it first goes through a stem module and two stages to obtain rich feature, which is next shared by $m$ branches to predict the AU occurrence probability $\hat{p}_{i}^{(j)}$, respectively. Each AU branch applies constraint to the self-attention \highlight{weight} distribution of an intermediate block in the third stage via an attention regression loss $\mathcal{L}_a$, and then uses causal intervention to deconfound the sample confounder in AU feature $\mathbf{f}_i^{(j)}$ under the guidance of AU detection loss $\mathcal{L}_u$. The formula $c'\times l'\times l'$ attached to each module denotes the size of its output, and $\times n$ denotes replicating the structure for $n$ times. ``$\star$'' and $+$ denote element-wise multiplication and element-wise addition, respectively. }
\label{fig:AC2D_framework}
\end{figure*}

\section{Related Work}
We review the previous works that are closely related to our method, including facial AU detection with self-attention and facial AU detection with causal inference. %causal inference in computer vision.

\subsection{Facial AU Detection with Self-Attention}

Traditional methods for AU detection often rely on handcrafted features and conventional machine learning algorithms~\citep{valstar2006fully,li2013simultaneous,zhao2016joint}, which have limitations in extracting powerful features and capturing complex dependencies. % and handling subtle, dynamic, and diverse AUs.
In the past decade, researchers have started exploring the use of deep learning techniques for AU detection, motivated by the great success of deep learning in computer vision. These methods often use convolutional neural networks (CNNs) to extract local region features~\citep{li2018eac,shao2021jaa}, use recurrent neural networks (RNNs) or long short-term memory (LSTM) networks to capture temporal dependencies~\citep{he2017multi,chu2017learning}, or use graph neural networks (GNNs) to model correlations among AUs as well as temporal dependencies~\citep{li2019semantic,song2021uncertain,shao2023facial}. However, due to the difficulty of handling subtle, dynamic, and diverse AUs, AU detection is still a challenging problem.
% To address these challenges,

In recent years, transformer with self-attention mechanism~\citep{vaswani_attention_2017} is introduced to the field of computer vision~\citep{dosovitskiy2021image}, and has gained increasing attention. Inspired by its global dependency modeling ability, \cite{jacob2021facial} and \cite{wang2022action} input AU features to a transformer for relational modeling among AUs, in which the AU features are extracted by CNNs. Since vision transformer (ViT)~\citep{dosovitskiy2021image} also has a strong feature learning ability, such use of CNN can be avoided. However, integrating the local feature extraction advantage of CNN and the global relational modeling advantage of vanilla transformer~\citep{vaswani_attention_2017} into a new ViT has been rarely explored in the AU detection field. In this work, we simplify a powerful ViT of ResTv2~\citep{zhang2022rest} as the backbone of our AU detection framework, which is effective in capturing local information and modeling global correlation, and is computationally efficient in self-attention.
Besides, we innovatively propose to adaptively constrain the self-attention \highlight{weight} distribution, which can combine the merits of both prior knowledge and self-attention learning.
% One prominent approach is the utilization of self-attention mechanisms. Self-attention allows the model to focus on different parts of the face, attending to the most informative regions for AU prediction.

% \subsection{Causal Inference in Computer Vision}
\subsection{Facial AU Detection with Causal Inference}

The main goal of causal inference~\citep{pearl2000models,rubin2005causal} is to learn the causal effect so as to eliminate spurious correlations~\citep{liu2022show} and disentangle desired effects~\citep{besserve2020counterfactuals}. It has significantly improved the performance of many computer vision tasks such as image classification~\citep{lopez2017discovering}, semantic segmentation~\citep{yue2020interventional}, and visual dialog~\citep{qi2020two}. For instance, \cite{zhang2020causal} introduced Pearl’s structural causal model~\citep{pearl2000models} to analyze the causalities among image, context \highlight{prior, image-specific context representation}, and class label, and then used the backdoor adjustment~\citep{pearl2016causal} to remove the confounding effect. %\highlight{In contrast,}

Recently, \highlight{inspired by \cite{zhang2020causal}'s work, }
\cite{chen2022causal} firstly introduced causal inference to the AU detection community \highlight{by formulating the causalities among image, subject, latent AU semantic relation, and AU label}, and removed the bias caused by subject confounder. It adopts a common causal intervention module for all AUs, and relies on the identity annotations of training data. However, the bias resulted from the variations across samples is unresolved, since %a same
\highlight{the} same subject may still present %a same
\highlight{the} same AU with different appearances and dependencies in different time or scenarios. In contrast, our method deconfounds the sample confounder in each AU branch, without the dependence on identity annotations. Besides, the deconfounding of subject confounder can be treated as a special case of our method. To our knowledge, our method is the second \highlight{work of} exploring causal inference based AU detection.

% \section{AC$^{2}$D for Facial AU Detection}
\section{Methodology}
\subsection{Overview}

Given the $i$-th facial image with size $3\times l\times l$ in the dataset, our main goal is to predict its AU occurrence probabilities $\hat{\mathbf{p}}_{i}=(\hat{p}_{i}^{(1)}, \cdots, \hat{p}_{i}^{(m)})$, where $m$ is the number of AUs. The structure of our AC$^{2}$D framework is shown in Fig.~\ref{fig:AC2D_framework}. To make it appropriate for AU detection, we simplify the structure of ResTv2~\citep{zhang2022rest} as our backbone. Specifically, a stem module is first used to capture low-level feature with both the height and width \highlight{dimensions} shrunk. The two stages with each composed of a patch embedding~\highlight{\citep{dosovitskiy2021image}} and multiple blocks are next adopted to extract rich feature \highlight{with abundant facial related information}. Each block consists of an efficient multi-head self-attention v2 (EMSAv2)~\citep{zhang2022rest} and a multilayer perceptron (MLP). EMSAv2 simplifies the structure of EMSA~\citep{zhang2021rest} by removing the multi-head interaction module, and adds an upsample module
including a depth-wise convolution and a pixel shuffle to reconstruct the lost medium- and high-frequency information during downsampling process.

Then, each AU has an independent branch to predict its occurrence probability $\hat{p}_{i}^{(j)}$ by feeding the rich feature, which contains the third stage and a causal intervention module. To exploit the prior knowledge about AU locations, we encourage the average self-attention \highlight{weight} distribution \highlight{$\mathbf{A}_i^{avg(j)}$} of the $(n_3-1)$-th block in the third stage to close to the predefined ground-truth attention \highlight{$\mathbf{A}_i^{gt(j)}$} via an attention regression loss $\mathcal{L}_a$, in which each individual self-attention channel in $\mathbf{A}_i^{(j)}$ is also adaptively learned under the supervision of the AU detection loss $\mathcal{L}_u$. \highlight{After adding a layer normalization layer and an adaptive average pooling layer to the end of the third stage, we obtain the feature $\mathbf{f}_i^{(j)}$ of the $j$-th AU.} Besides, to remove the bias caused by inherent sample characteristics, %and specific image scenes,
% variations across samples,
causal intervention is adopted to deconfound the sample confounder in AU feature $\mathbf{f}_i^{(j)}$ via backdoor adjustment~\citep{pearl2016causal}. Our framework including self-attention constraining and causal intervention is end-to-end trainable, in which the rich feature shared by all AUs can exploit common patterns, and the separate branch for each AU is beneficial for modeling AU-specific \highlight{causal} characteristics.

\begin{figure}
\centering\includegraphics[width=\linewidth]{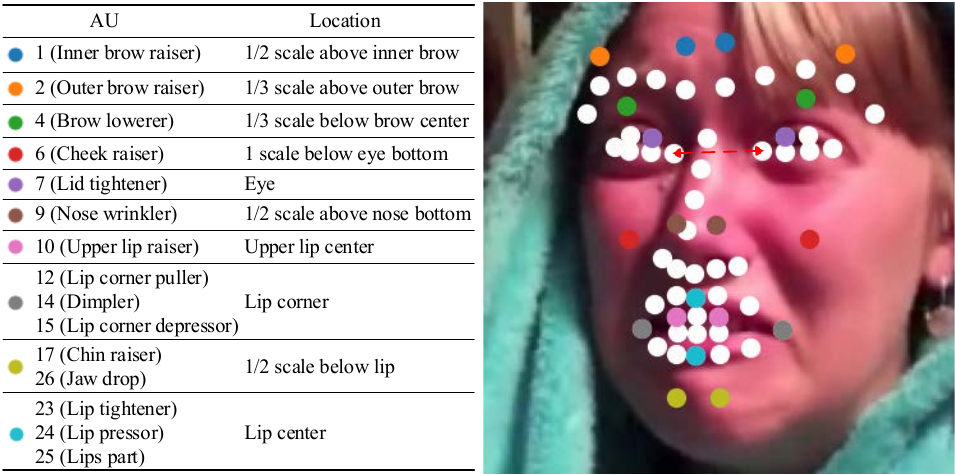}
\caption{Definition to the locations of AU sub-centers, which is applicable to an aligned face with eye centers on the same horizontal line~\citep{li2018eac,shao2021jaa}. Each AU has two sub-centers specified by \highlight{two} facial landmarks \highlight{due to facial symmetry}. The red dotted line denotes the distance between two inner eye corners, i.e. ``scale''.}
\label{fig:au_landmark_location}
\end{figure}

\subsection{Adaptive Constraining on Self-Attention}

\highlight{Self-attention~\citep{vaswani_attention_2017} is known as a powerful long-range relational modeling ability, but has limitations in extracting local features. To resolve this issue, we propose to constrain the self-attention by exploiting prior knowledge about AU locations.} The scaled dot-product attention weight is defined as
\begin{equation}
    \mathbf{A}_i=Softmax(\frac{\mathbf{Q}_i\mathbf{K}_i^T}{\sqrt{d'}}),
\end{equation}
where $\mathbf{Q}_i\in \mathbb{R}^{k'\times n'\times d'}$, $\mathbf{K}_i\in \mathbb{R}^{k'\times n'\times d'}$, $\mathbf{A}_i\in \mathbb{R}^{k'\times n'\times n'}$, and $Softmax(\cdot)$ denotes a Softmax function. \highlight{$Softmax(\cdot)$ is computed along the last dimension so that all values along the last dimension of $\mathbf{A}_i$ sum to $1$. \textit{In this case, each channel in the last dimension of $\mathbf{A}_i$ conforms to a distribution, and we call $\mathbf{A}_i$ as self-attention weight distribution.}} As illustrated in Fig.~\ref{fig:AC2D_framework}, the self-attention~\citep{zhang2022rest} process is defined as
\begin{equation}
    SA(\mathbf{Q}_i, \mathbf{K}_i, \mathbf{V}_i)=\mathbf{A}_i\mathbf{V}_i+UP(\mathbf{V}_i),
\end{equation}
where $\mathbf{V}_i\in \mathbb{R}^{k'\times n'\times d'}$, and $UP(\cdot)$ denotes the operation of the upsample module. \highlight{In our AC$^{2}$D network, we conduct self-attention constraining in each AU branch, and we denote the self-attention weight distribution in the $j$-th AU branch as $\mathbf{A}_i^{(j)}$.}

As shown in Fig.~\ref{fig:au_landmark_location},  the locations of AUs can be specified by correlated facial landmarks~\citep{li2018eac,shao2021jaa}, in which each AU has two sub-centers. By exploiting this prior knowledge, we can predefine the ground-truth attention $\mathbf{A}_i^{gt(j)}\in \mathbb{R}^{l/8\times l/8}$ for the $j$-th AU. We first generate the predefined attention \highlight{$\widetilde{\mathbf{A}}_i^{gt(j),1}$} with regard to one sub-center \highlight{$(\bar{a}_{i}^{gt(j),1},\bar{b}_{i}^{gt(j),1})$} via a Gaussian distribution with standard deviation $\delta$, in which the value at location $(a,b)$ is defined as
\begin{equation}
\label{eq:pre_att}
    \widetilde{A}_{iab}^{gt(j),1} = \exp(-\frac{(a-\bar{a}_{i}^{gt(j),1})^2+(b-\bar{b}_{i}^{gt(j),1})^2}{2\delta^2}).
\end{equation}
Then, we combine the predefined attentions \highlight{$\widetilde{\mathbf{A}}_i^{gt(j),1}$} and \highlight{$\widetilde{\mathbf{A}}_i^{gt(j),2}$} of both sub-centers by choosing the larger value at each location $(a,b)$:
\begin{equation}
\label{eq:combine_pre_att}
    \widetilde{A}_{iab}^{gt(j)} = \max(\widetilde{A}_{iab}^{gt(j),1}, \widetilde{A}_{iab}^{gt(j),2})\in (0,1].
\end{equation}
% where $A_{iab}^{gt(j)}$.
Finally, we normalize \highlight{$\widetilde{\mathbf{A}}_{i}^{gt(j)}$} so as to conform to a distribution with all values summing to 1:
\begin{equation}
\label{eq:normalize_att}
    A_{iab}^{gt(j)}=\widetilde{A}_{iab}^{gt(j)}/\sum_{s=1}^{l/8}\sum_{t=1}^{l/8}\widetilde{A}_{ist}^{gt(j)},
\end{equation}
where a lower value is assigned to a location farther away from both AU sub-centers in \highlight{$\mathbf{A}_i^{gt(j)}$}.

Since self-attention captures the characteristics of facial AUs in an AU detection network, the scaled dot-product attention weight $\mathbf{A}_i^{(j)}\in \mathbb{R}^{k\times (l/8\times l/8)\times (l/8\times l/8)}$ in the $(n_3-1)$-th block of the third stage can be regarded as multiple spatial attentions by reshaping to be the size of $(k\times l/8\times l/8)\times (l/8\times l/8)$. To reserve enough space for automatic self-attention learning, we choose to constrain the average self-attention \highlight{weight} distribution. The average of $\mathbf{A}_i^{(j)}$ over $k\times l/8\times l/8$ channels is calculated as
\begin{equation}
    \mathbf{A}_i^{avg(j)}=\frac{1}{k\times l/8\times l/8}\sum_{s=1}^{k\times l/8\times l/8}\mathbf{A}_{is}^{(j)},
\end{equation}
where \highlight{$\mathbf{A}_{is}^{(j)}$ and $\mathbf{A}_i^{avg(j)}$ both have} the same size $l/8\times l/8$ as \highlight{$\mathbf{A}_i^{gt(j)}$, and also both conform to a distribution with all values summing to 1}. We adopt a Kullback-Leibler (KL) divergence~\citep{kullback1951information} loss to encourage \highlight{$\mathbf{A}_i^{avg(j)}$} to close to \highlight{$\mathbf{A}_i^{gt(j)}$}:
\begin{equation}
\label{eq:L_attention}
\begin{aligned}
\mathcal{L}_{a} =& \frac{1}{m(l/8\times l/8)}\sum_{j=1}^m\sum_{s=1}^{l/8}\sum_{t=1}^{l/8} (\\
&A_{ist}^{gt(j)}\log A_{ist}^{gt(j)} - A_{ist}^{gt(j)}\log A_{ist}^{avg(j)}),
\end{aligned}
\end{equation}
where KL divergence measures the differences between two distributions, and \highlight{constrained $\mathbf{A}_i^{(j)}$ has numerous attention map channels to extract} % this loss is beneficial for extracting
AU-related region features. Under the constraint on self-attention \highlight{weight} distribution and the guidance from AU detection loss, self-attention is adaptively constrained and can more accurately model the characteristics of AUs, \highlight{in which local information related to each AU is captured while global relational information is still modeled.}

\begin{figure}
\centering\includegraphics[width=\linewidth]{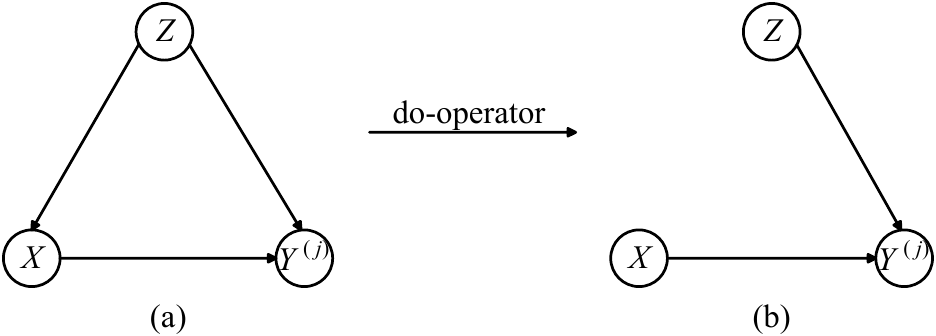}
\caption{Illustration of our causal diagram for each AU. (a) The conventional likelihood $P(Y^{(j)}|X)$. (b) The likelihood $P(Y^{(j)}|do(X))$ after causal intervention.}
\label{fig:casual_diagram}
\end{figure}

\subsection{Causal Deconfounding of Sample Confounder}

We adopt Pearl’s structural causal model~\citep{pearl2000models} to analyze the causal relationships. Fig.~\ref{fig:casual_diagram}(a) shows the causal diagram among facial image $X$, sample \highlight{confounder (also known as sample characteristics)} $Z$, and AU occurrence probability $Y^{(j)}$ for the $j$-th AU. The direction of an edge represents the causal relationship. For example, $X\rightarrow Y^{(j)}$ denotes that $X$ is the cause and $Y^{(j)}$ is the effect. The causal relationships are elaborated below:
\begin{itemize}
    \item $X\rightarrow Y^{(j)}$. The AU occurrence probability predicted by an AU detection network relies on the input facial image, in which this causal relationship is intended to learn by the network.
    \item $Z\rightarrow X$. %The appearance of an AU is determined by , as well as
    The time and scenario recording the sample \highlight{determines image background, image illumination, and image quality, and influences the} emotion of the subject corresponding to the sample. \highlight{Besides,} the custom of expressing emotion of the subject determines the appearances of AUs \highlight{in the facial image.}

    \item $Z\rightarrow Y^{(j)}$.
    % Since the  of  are
    \highlight{Besides} the sample characteristics embedded in the facial image, \highlight{the sample characteristics in terms of outside scenario like certain social interaction can influence the correlations among AUs including co-occurrences and exclusions.
    % and will be embedded in the trained network
    Therefore}, we have the causal link from $Z$ to $Y^{(j)}$.
\end{itemize}

To eliminate the effect brought by confounder $Z$ so that the trained network predicts \highlight{$Y^{(j)}$} %$Z\rightarrow Y^{(j)}$
only based on $X$, we block the backdoor path between $Z$ and $X$ via a do-operator, as shown in Fig.~\ref{fig:casual_diagram}(b). In this way, we learn an AU detection network by solving $P(Y^{(j)}|do(X))$ instead of $P(Y^{(j)}|X)$. A straightforward solution to deconfound the sample confounder is to collect all the sample images so that $P(Y^{(j)}|X)$ equals to $P(Y^{(j)}|do(X))$. Considering such way is not practical due to the infinite number of samples, we apply the backdoor adjustment~\citep{pearl2016causal} technique. Particularly, we estimate the causal effect for each sample in the training set and then compute the average causal effect:
\begin{equation}
\label{eq:doX_1}
    P(Y^{(j)}|do(X)) =\sum_{z}P(Y^{(j)}|X, Z=z)P(Z=z),
\end{equation}
where $X$ is no longer dependent on $Z$, and $X$ \highlight{considers} every sample $z$ into the prediction of $Y^{(j)}$ based on the ratio of $z$ in the whole.
%training set.%has a fair chance to

As illustrated in Fig.~\ref{fig:AC2D_framework}, the learned feature $\mathbf{f}_i^{(j)}\in \mathbb{R}^{c}$ of the $j$-th AU for the $i$-th input image $X$ is fed into a causal intervention module.
In Eq.~\eqref{eq:doX_1}, each pair of $X$ and $z$ is required. To reduce the computational costs, we use normalized weighted geometric mean (NWGM)~\highlight{\citep{xu2015show}} technique to approximate Eq.~\eqref{eq:doX_1}:
\begin{equation}
    P(Y^{(j)}|do(X)) \approx P(Y^{(j)}|X, Z=\sum_{z} zP(z)).
\end{equation}
This conditional probability can be implemented as a linear model~\citep{wang2020visual}:
\begin{equation}
\label{eq:causal_linear}
    P(Y^{(j)}|do(X))=\mathbf{W}_X^{(j)}\mathbf{f}_i^{(j)}+\mathbf{W}_Z^{(j)}\mathbb{E}_z[g(z)],
\end{equation}
where \highlight{$\mathbb{E}_z[g(z)]$ is the approximation of sample confounder $Z$, AU feature $\mathbf{f}_i^{(j)}$ is extracted from the input image before causal intervention,} and $\mathbf{W}_X^{(j)}\in \mathbb{R}^{8c\times c}$ and $\mathbf{W}_Z^{(j)}\in \mathbb{R}^{8c\times c}$ are learnable parameters.  %\highlight{to combine AU feature and confounder}. The  %\highlight{Eq.~\eqref{eq:causal_linear} is reasonable since the effect $Y^{(j)}$ comes from both image $X$ and confounder $Z$.} for current input image

We formulate $\mathbb{E}_z[g(z)]$ as a weighted combination of all the sample prototypes $[z_1, z_2, \cdots, z_N]$\highlight{~\citep{wang2020visual}}:
\begin{equation}
\label{eq:gz_1}
    \mathbb{E}_z[g(z)]=\sum_{s=1}^{N}\alpha_s z_s P(z_s),
\end{equation}
where $N$ is the number of sample prototypes, and $\alpha_s$ is a coefficient for current AU feature $\mathbf{f}_i^{(j)}$. Since each sample prototype $z_s$ only has one image in the training set, we have $P(z_s)=\frac{1}{N}$, set $z_s$ as $\mathbf{f}_s^{(j)}$, and can set an equal coefficient $\alpha_s$ using scaled dot-product attention~\citep{vaswani_attention_2017} for all sample prototypes:
\begin{subequations}
\begin{equation}
    \bar{\mathbf{f}}^{(j)} = \frac{1}{N}\sum_{s=1}^{N}\mathbf{f}_s^{(j)},
\end{equation}
\begin{equation}
    \mathbf{Q}_i^{o(j)} = \mathbf{W}_Q^{(j)} \mathbf{f}_i^{(j)},
\end{equation}
\begin{equation}
    \mathbf{K}^{o(j)} = \mathbf{W}_K^{(j)} \bar{\mathbf{f}}^{(j)},
\end{equation}
\begin{equation}
    \alpha_s=Softmax(\frac{\mathbf{Q}_i^{o(j)}\mathbf{K}^{o(j)T}}{\sqrt{8c}}),
\end{equation}
\end{subequations}
where $\mathbf{W}_Q^{(j)}\in \mathbb{R}^{8c\times c}$ and $\mathbf{W}_K^{(j)}\in \mathbb{R}^{8c\times c}$ are learnable parameters, and $\mathbf{f}_s^{(j)}$ is updated in each training epoch. In this way, Eq.~\eqref{eq:gz_1} can be rewritten as
\begin{equation}
\label{eq:gz_2}
    \mathbb{E}_z[g(z)]=Softmax(\frac{\mathbf{Q}_i^{o(j)}\mathbf{K}^{o(j)T}}{\sqrt{8c}})\bar{\mathbf{f}}^{(j)}.
\end{equation}

\highlight{In Eq.~\eqref{eq:gz_2}, the AU feature $\mathbf{f}_i^{(j)}$ of the $i$-th sample prototype and the average AU feature $\bar{\mathbf{f}}^{(j)}$ over all sample prototypes are interacted in a self-attention structure to approximate the sample confounder. Besides, the computation participation of $\bar{\mathbf{f}}^{(j)}$ in Eq.~\eqref{eq:gz_2} is reasonable since samples from the training set often have similar or relevant outside scenarios. In Eq.~\eqref{eq:causal_linear}, this causal intervention process can be seen as learning sample-deconfounded AU feature.} Finally, the predicted AU occurrence probability $\hat{p}_{i}^{(j)}$ can be obtained by adding a one-dimensional linear layer with a Sigmoid function. In our AC$^{2}$D, we deconfound the sample confounder for each AU separately, which contributes to modeling AU-specific \highlight{causal} patterns.

We use an AU detection loss with weighting strategy~\citep{shao2023facial}:
\begin{equation}
\label{eq:L_u}
\mathcal{L}_{u}\!=\!-\!\sum_{j=1}^m w_j [v_j p_i^{(j)}\! \log \hat{p}_{i}^{(j)} + (1-p_i^{(j)}) \log (1-\hat{p}_{i}^{(j)})],
\end{equation}
where $w_j=\frac{N}{N^{occ(j)}}/\sum_{s=1}^{m}\frac{N}{N^{occ(s)}}$ is the weight of the $j$-th AU, $v_j=\frac{N-N^{occ(j)}}{N^{occ(j)}}$ is the weight for occurrence of the $j$-th AU, and $p_i^{(j)}$ denotes the ground-truth occurrence probability of the $j$-th AU. $N^{occ(j)}$ is the number of samples occurring the $j$-th AU in the training set, and the occurrence rate of the $j$-th AU can be computed as $N^{occ(j)}/{N}$. This weighting strategy is beneficial for suppressing two types of data imbalance problems: different AUs have different occurrence rates, and occurrence rate is often lower than non-occurrence rate for an AU.

By incorporating Eqs.~\eqref{eq:L_attention} and~\eqref{eq:L_u}, we obtain the complete loss:
\begin{equation}
\label{eq:L_AA}
\mathcal{L} = \mathcal{L}_{u} + \lambda_a \mathcal{L}_{a},
\end{equation}
where $\lambda_a$ controls the importance of $\mathcal{L}_{a}$. In our framework, adaptive constraining on self-attention \highlight{weight} distribution and causal deconfounding of sample confounder are simultaneously optimized, which jointly contribute to AU detection.

\section{Experiments}

\subsection{Datasets and Settings}

\subsubsection{Datasets}

Our AC$^2$D is evaluated on \highlight{five} benchmark datasets, in terms of BP4D~\citep{zhang2014bp4d}, DISFA~\citep{mavadati2013disfa}, GFT~\citep{girard2017sayette}, and \highlight{BP4D+~\citep{zhang2016multimodal}} in constrained scenarios, and Aff-Wild2~\citep{kollias2019expression,kollias2021analysing} in unconstrained scenarios.
\begin{itemize}
    \item\textbf{BP4D} includes $23$ females and $18$ males, each of which participates in $8$ sessions. There are about $140,000$ frames annotated by AU labels of occurrence or non-occurrence. Each frame is also annotated by $49$ facial landmarks. Following the settings in~\cite{zhao2016deep,li2018eac,shao2021jaa}, we evaluate on $12$ AUs (1, 2, 4, 6, 7, 10, 12, 14, 15, 17, 23, and 24) using subject exclusive 3-fold cross-validation, in which two folds are used for training and the remaining one is used for testing.
    \item\textbf{DISFA} contains $27$ videos captured from $12$ females and $15$ males, each of which includes $4,845$ frames. Each frame is annotated by AU intensities on a six-point ordinal scale from $0$ to $5$, as well as $66$ facial landmarks. We use the settings in~\cite{zhao2016deep,li2018eac,shao2021jaa} by treating AU intensities equal or greater than $2$ as occurrence and otherwise treating as non-occurrence. We also adopt the subject exclusive 3-fold cross-validation, and evaluate on $8$ AUs: 1, 2, 4, 6, 9, 12, 25, and 26.
    \item\textbf{GFT} includes $96$ subjects from $32$ three-subject groups in unscripted talks. Each subject is captured by a video, in which most frames exhibit moderate out-of-plane poses. Each frame is annotated by $10$ AUs (1, 2, 4, 6, 10, 12, 14, 15, 23, and 24), as well as $49$ facial landmarks. Following the official training/testing partitions~\citep{girard2017sayette}, we utilize $78$ subjects with about $108,000$ frames for training, and utilize $18$ subjects with about $24,600$ frames for testing.
    \highlight{\item\textbf{BP4D+} contains $82$ female and $58$ male subjects, and each subject is involved in $10$ sessions. This dataset has larger scale and diversity than BP4D~\citep{zhang2014bp4d} dataset. There are $4$ sessions including totally $197,875$ frames with AU annotations, in which each frame is also annotated by $49$ facial landmarks. We use the cross-dataset evaluation settings in~\cite{shao2019facial,shao2021jaa} by training on the whole BP4D dataset ($41$ subjects with $12$ AUs) and testing on the whole BP4D+ dataset.}
    \item\textbf{Aff-Wild2} is a large-scale in-the-wild dataset collected from YouTube. It contains a training set including $305$ videos with about $1,390,000$ frames, and a validation set including $105$ videos with about $440,000$ frames. Each frame is annotated by $12$ AUs (1, 2, 4, 6, 7, 10, 12, 15, 23, 24, 25, and 26), and shows diverse variations in ages, ethnicities, professions, emotions, poses, illumination, or occlusions. We use $68$ facial landmark annotations on each frame provided by~\cite{shao2023facial}, and follow its setting with training on the training set and testing on the validation set.
\end{itemize}

\begin{table*}
\centering\caption{F1-frame results for $12$ AUs on BP4D~\citep{zhang2014bp4d}. The results of previous methods are reported in their original papers.}
\label{tab:comp_f1_bp4d}
\begin{tabular}{c|*{12}{c}|c}
\toprule
AU &1 &2 &4 &6 &7 &10 &12 &14 &15 &17 &23 &24 &\textbf{Avg}\\
\midrule
DRML~\citep{zhao2016deep} &36.4 &41.8 &43.0 &55.0 &67.0 &66.3 &65.8 &54.1 &33.2 &48.0 &31.7 &30.0 &48.3\\
EAC-Net~\citep{li2018eac} &39.0 &35.2 &48.6 &76.1 &72.9 &81.9 &86.2 &58.8 &37.5 &59.1 &35.9 &35.8 &55.9\\
DSIN~\citep{corneanu2018deep}  &51.7 &40.4 &56.0 &76.1 &73.5 &79.9 &85.4 &62.7 &37.3 &62.9 &38.8 &41.6 &58.9\\
CMS~\citep{sankaran2019representation} &49.1 &44.1 &50.3 &79.2 &74.7 &80.9 &88.3 &63.9 &44.4 &60.3 &41.4 &51.2 &60.6\\
LP-Net~\citep{niu2019local} &43.4 &38.0 &54.2 &77.1 &76.7 &83.8 &87.2 &63.3 &45.3 &60.5 &48.1 &54.2 &61.0\\
SRERL~\citep{li2019semantic} &46.9 &45.3 &55.6 &77.1 &78.4 &83.5 &87.6 &60.6 &52.2 &63.9 &47.1 &53.3 &62.9\\
ARL~\citep{shao2019facial} &45.8 &39.8 &55.1 &75.7 &77.2 &82.3 &86.6 &58.8 &47.6 &62.1 &47.4 &55.4 &61.1\\
AU R-CNN~\citep{ma2019r} &50.2 &43.7 &57.0 &78.5 &78.5 &82.6 &87.0 &\textbf{67.7} &49.1 &62.4 &50.4 &49.3 &63.0\\
AU-GCN~\citep{liu2020relation} &46.8 &38.5 &60.1 &\textbf{80.1} &79.5 &84.8 &88.0 &67.3 &52.0 &63.2 &40.9 &52.8 &62.8\\
JÂA-Net~\citep{shao2021jaa}  &53.8 &47.8 &58.2 &78.5 &75.8 &82.7 &88.2 &63.7 &43.3 &61.8 &45.6 &49.9 &62.4\\
UGN-B~\citep{song2021uncertain} &54.2 &46.4 &56.8 &76.2 &76.7 &82.4 &86.1 &64.7 &51.2 &63.1 &48.5 &53.6 &63.3\\
HMP-PS~\citep{song2021hybrid} &53.1 &46.1 &56.0 &76.5 &76.9 &82.1 &86.4 &64.8 &51.5 &63.0 &49.9 &54.5 &63.4\\
\cite{jacob2021facial} &51.7 &49.3 &61.0 &77.8 &79.5 &82.9 &86.3 &67.6 &51.9 &63.0 &43.7 &56.3 &64.2\\
% \highlight{RTATL~\citep{yan2021self}} &\textbf{57.1} &\textbf{49.7} &60.5 &77.9 &76.1 &84.4 &87.2 &64.3 &53.5 &\textbf{67.0} &48.9 &48.6 &64.6\\
% \highlight{Li et al.~\citep{li2021integrating}} &54.0 &46.0 &55.7 &79.4 &78.8 &84.5 &87.0 &67.0 &55.6 &63.1 &50.7 &55.3 &\textbf{64.8}\\
AAR~\citep{shao2023facial} &53.2 &47.7 &56.7 &75.9 &79.1 &82.9 &88.6 &60.5 &51.5 &61.9 &51.0 &\textbf{56.8} &63.8\\
CISNet~\citep{chen2022causal} &54.8 &48.3 &57.2 &76.2 &76.5 &\textbf{85.2} &87.2 &66.2 &50.9 &65.0 &47.7 &56.5 &64.3\\
\cite{chang2022knowledge} &53.3 &47.4 &56.2 &79.4 &\textbf{80.7} &85.1 &\textbf{89.0} &67.4 &\textbf{55.9} &61.9 &48.5 &49.0 &64.5\\
\highlight{AUNet~\citep{yang2023toward}} &\textbf{58.0} &48.2 &\textbf{62.4} &76.4 &77.5 &83.4 &88.5 &63.3 &52.0 &\textbf{65.5} &\textbf{52.1} &52.3 &\textbf{65.0}\\
\textbf{AC$^{2}$D} &54.2&\textbf{54.7}&56.5&77.0&76.2&84.0&\textbf{89.0}&63.6&54.8&63.6&46.5&54.8&64.6\\
\bottomrule
\end{tabular}
\end{table*}

\begin{table*}
\centering\caption{F1-frame results for $8$ AUs on DISFA~\citep{mavadati2013disfa}.}
\label{tab:comp_f1_disfa}
\setlength\tabcolsep{11.4pt}
\begin{tabular}{c|*{8}{c}|c}
\toprule
AU &1 &2 &4 &6 &9 &12 &25 &26 &\textbf{Avg}\\
\midrule
DRML~\citep{zhao2016deep} &17.3 &17.7 &37.4 &29.0 &10.7 &37.7 &38.5 &20.1 &26.7\\
EAC-Net~\citep{li2018eac} &41.5 &26.4 &66.4 &50.7 &8.5 &\textbf{89.3} &88.9 &15.6 &48.5\\
DSIN~\citep{corneanu2018deep}  &42.4 &39.0 &68.4 &28.6 &46.8 &70.8 &90.4 &42.2 &53.6\\
CMS~\citep{sankaran2019representation} &40.2 &44.3 &53.2 &57.1 &50.3 &73.5 &81.1 &59.7 &57.4\\
LP-Net~\citep{niu2019local} &29.9 &24.7 &72.7 &46.8 &49.6 &72.9 &93.8 &65.0 &56.9\\
SRERL~\citep{li2019semantic} &45.7 &47.8 &59.6 &47.1 &45.6 &73.5 &84.3 &43.6 &55.9\\
ARL~\citep{shao2019facial} &43.9 &42.1 &63.6 &41.8 &40.0 &76.2 &95.2 &66.8 &58.7\\
AU R-CNN~\citep{ma2019r} &32.1 &25.9 &59.8 &55.3 &39.8 &67.7 &77.4 &52.6 &51.3\\
AU-GCN~\citep{liu2020relation} &32.3 &19.5 &55.7 &\textbf{57.9} &\textbf{61.4} &62.7 &90.9 &60.0 &55.0\\
JÂA-Net~\citep{shao2021jaa}  &\textbf{62.4} &\textbf{60.7} &67.1 &41.1 &45.1 &73.5 &90.9 &67.4 &63.5\\
UGN-B~\citep{song2021uncertain} &43.3 &48.1 &63.4 &49.5 &48.2 &72.9 &90.8 &59.0 &60.0\\
HMP-PS~\citep{song2021hybrid} &38.0 &45.9 &65.2 &50.9 &50.8 &76.0 &93.3 &67.6 &61.0\\
\cite{jacob2021facial}&46.1 &48.6 &72.8 &56.7 &50.0 &72.1 &90.8 &55.4 &61.5\\
% \highlight{RTATL~\cite{yan2021self}} &57.8 &52.8 &70.8 &53.2 &52.7 &74.5 &91.5 &51.9 &63.1\\
% \highlight{Li et al.~\cite{li2021integrating}} &47.5 &53.3 &64.4 &51.8 &44.4 &74.7 &92.1 &60.7 &61.1\\
AAR~\citep{shao2023facial} &\textbf{62.4} &53.6 &71.5 &39.0 &48.8 &76.1 &91.3 &\textbf{70.6} &64.2\\
CISNet~\citep{chen2022causal} &48.8 &50.4 &\textbf{78.9} &51.9 &47.1 &80.1 &\textbf{95.4} &65.0 &64.7\\
\cite{chang2022knowledge} &60.4 &59.2 &67.5 &52.7 &51.5 &76.1 &91.3 &57.7 &64.5\\
\highlight{AUNet~\citep{yang2023toward}} &60.3 &59.1 &69.8 &48.4 &53.0 &79.7 &93.5 &64.7 &\textbf{66.1}\\
\textbf{AC$^{2}$D} &57.8&59.2&70.1&50.1&54.4&75.1&90.3&66.2&65.4\\
\bottomrule
\end{tabular}
\end{table*}

\subsubsection{Implementation Details}

Each face image is aligned to $3\times 200\times 200$ using similarity transformation via fitting facial landmarks. To augment the training data, the image is randomly cropped to $3\times 176\times 176$ as the input of our network, and is further conducted with random mirroring and random color jittering in terms of contrast and brightness. The dimension parameters $c$ and $k$, the crop size $l$, the structure parameters $n_1$, $n_2$, and $n_3$, and the standard deviation $\delta$ are set to $64$, $4$, $176$, $1$, $6$, $3$, and $3$, respectively. The number of AUs $m$ is $12$, $8$, $10$, \highlight{$12$}, and $12$ in BP4D, DISFA, GFT, \highlight{BP4D+}, and Aff-Wild2, respectively. \highlight{To choose an appropriate value for the trade-off parameter $\lambda_a$, we select multiple small sets from the training set of Aff-Wild2 as validation sets. When evaluating on each small validation set, we train AC$^2$D on the training set excluding the current validation set. $\lambda_a$ is chosen as $1.28\times 10^4$ for the overall best performance on the validation sets, and is fixed for other datasets.}

Our AC$^2$D \highlight{uses a simplified structure of the tiny version of ResTv2~\citep{zhang2022rest}, and} is implemented using PyTorch~\citep{paszke2019pytorch}. Similar to the settings in ResTv2, we train AC$^2$D for up to $20$ epochs using AdamW~\citep{loshchilov2017decoupled} optimizer, with a cosine decay learning rate scheduler and $1$ epoch for linear warm-up, an initial learning rate of $2\times 10^{-3}/256$ multiplying the mini-batch size, a weight
decay of $0.05$, and gradient clipping~\citep{zhang2019gradient} with a max norm of $3.0$. Following the previous works\citep{zhao2016deep,li2018eac,shao2021jaa}, our AC$^2$D model trained on DISFA is initialized using the parameters of our well-trained model on BP4D.

\subsubsection{Evaluation Metrics}

We report a popular metric of frame-based F1-score (F1-frame) in AU detection: $F1=2PR/(P+R)$, where $P$ and $R$ mean precision and recall, respectively. We also report the average F1-frame over all AUs, abbreviated as Avg. In the following sections, we show all the F1-frame results in percentage with ``$\%$'' omitted.

\subsection{Comparison with State-of-the-Art Methods}

Our AC$^{2}$D is compared against state-of-the-art AU detection methods under the same evaluation setting, including LSVM~\citep{fan2008liblinear}, AlexNet~\citep{krizhevsky2012imagenet}, DRML~\citep{zhao2016deep}, EAC-Net~\citep{li2018eac}, DSIN~\citep{corneanu2018deep}, CMS~\citep{sankaran2019representation}, LP-Net~\citep{niu2019local}, SRERL~\citep{li2019semantic}, ARL~\citep{shao2019facial}, AU R-CNN~\citep{ma2019r}, TCAE~\citep{li2019self-supervised}, AU-GCN~\citep{liu2020relation}, \cite{ertugrul2020crossing}, JÂA-Net~\citep{shao2021jaa}, UGN-B~\citep{song2021uncertain}, HMP-PS~\citep{song2021hybrid}, \cite{zhang2021prior}, \cite{jacob2021facial}, AAR~\citep{shao2023facial}, CISNet~\citep{chen2022causal}, \cite{chang2022knowledge}, \highlight{and AUNet~\citep{yang2023toward}.}

\highlight{Note that AAR and AUNet use temporal information, and other methods process a single image at a time without utilizing temporal information. Besides,} most of these previous methods use outside training data, while our approach only uses training data from the benchmark dataset. In particular, EAC-Net, SRERL, AU R-CNN, UGN-B, HMP-PS, \cite{jacob2021facial}, and \cite{chang2022knowledge} fine-tune pre-trained VGG~\citep{simonyan2014very}, ResNet~\citep{he2016deep}, or InceptionV3~\citep{szegedy2016rethinking} models, \highlight{AUNet uses a pretrained stacked hourglass network~\citep{newell2016stacked,toisoul2021estimation} and a pretrained variational autoencoder~\citep{diederik2014auto,luo2020shape},} CMS adopts outside thermal images, LP-Net pre-trains on a face recognition dataset, CISNet uses additional facial identity annotations, and \cite{zhang2021prior} utilizes BP4D~\citep{zhang2014bp4d} dataset when trained on Aff-Wild2~\citep{kollias2019expression,kollias2021analysing}.

\begin{table*}
\centering\caption{F1-frame results for $10$ AUs on GFT~\citep{girard2017sayette}. The results of LSVM~\citep{fan2008liblinear} and AlexNet~\citep{krizhevsky2012imagenet} are reported in~\cite{girard2017sayette}, and those of EAC-Net~\citep{li2018eac} and ARL~\citep{shao2019facial} are reported in~\cite{shao2021jaa}.}
\label{tab:comp_f1_gft}
\setlength\tabcolsep{7.7pt}
\begin{tabular}{c|*{10}{c}|c}
\toprule
AU&1&2&4&6&10&12&14&15&23&24&\textbf{Avg}\\
\midrule
LSVM~\citep{fan2008liblinear} &38 &32 &13 &67 &64
&78 &15 &29 &49 &44 &42.9\\
AlexNet~\citep{krizhevsky2012imagenet} &44 &46 &2 &73 &72 &82 &5 &19 &43 &42 &42.8\\
EAC-Net~\citep{li2018eac} &15.5 &56.6 &0.1 &81.0 &76.1 &84.0 &0.1 &38.5 &57.8 &\textbf{51.2} &46.1\\
TCAE~\citep{li2019self-supervised} &43.9 &49.5 &6.3
&71.0 &76.2 &79.5 &10.7 &28.5 &34.5 &41.7 &44.2\\
ARL~\citep{shao2019facial} &51.9 &45.9 &13.7 &79.2 &75.5 &82.8 &0.1 &44.9 &\textbf{59.2} &47.5 &50.1\\
~\cite{ertugrul2020crossing} &43.7 &44.9 &19.8 &74.6 &\textbf{76.5} &79.8 &50.0 &33.9 &16.8 &12.9 &45.3\\
JÂA-Net~\citep{shao2021jaa} &46.5 &49.3 &19.2 &79.0 &75.0 &84.8 &44.1 &33.5 &54.9 &50.7 &53.7\\
AAR~\citep{shao2023facial} &\textbf{66.3} &53.9 &23.7 &81.5 &73.6 &84.2 &43.8 &\textbf{53.8} &58.2 &46.5 &58.5\\
\textbf{AC$^2$D} &60.9&\textbf{58.2}&\textbf{24.4}&\textbf{83.3}&75.9&\textbf{87.4}&\textbf{56.4}&46.5&58.3&50.9&\textbf{60.2}\\
\bottomrule
\end{tabular}
\end{table*}

\begin{table*}
\centering\caption{\highlight{F1-frame results for $12$ AUs on BP4D+~\citep{zhang2016multimodal} in terms of cross-dataset evaluation. The results of EAC-Net~\citep{li2018eac} are reported in~\cite{shao2021jaa}.}}
\label{tab:bp4d_plus_AUocc}
\begin{tabular}{c|*{12}{c}|c}
\toprule
AU&1&2&4&6&7&10&12&14&15&17&23&24&\textbf{Avg}\\
\midrule
\highlight{EAC-Net~\citep{li2018eac}}&38.0 &\textbf{37.5} &\textbf{32.6} &82.0 &83.4 &87.1 &85.1 &62.1 &44.5 &43.6 &45.0 &32.8 &56.1\\
\highlight{ARL~\citep{shao2019facial}}&29.9 &33.1 &27.1 &81.5 &83.0 &84.8 &86.2 &59.7 &44.6 &43.7 &48.8 &32.3 &54.6\\
\highlight{JÂA-Net~\citep{shao2021jaa}}&39.7 &35.6 &30.7 &\textbf{82.4} &84.7 &88.8 &\textbf{87.0} &62.2 &38.9 &\textbf{46.4} &48.9 &\textbf{36.0} &56.8\\
\highlight{\textbf{AC$^2$D}}&\textbf{42.3}&35.4&26.7&80.7&\textbf{87.0}&\textbf{90.9}&85.8&\textbf{73.3}&\textbf{45.3}&43.4&\textbf{50.3}&29.0&\textbf{57.5}\\
\bottomrule
\end{tabular}
\end{table*}

\begin{table*}
\centering\caption{F1-frame results for $12$ AUs on Aff-Wild2~\citep{kollias2019expression,kollias2021analysing}. The results of EAC-Net~\citep{li2018eac}, ARL~\citep{shao2019facial}, and JÂA-Net~\citep{shao2021jaa} are reported in~\cite{shao2023facial}.}
\label{tab:comp_f1_affwild2}
\setlength\tabcolsep{6.3pt}
\begin{tabular}{c|*{12}{c}|c}
\toprule
AU &1 &2 &4 &6 &7 &10 &12 &15 &23 &24 &25 &26 &\textbf{Avg}\\
\midrule
EAC-Net~\citep{li2018eac} &49.6 &33.7 &55.6 &66.4 &82.3 &81.4 &76.9 &11.8 &12.5 &12.2 &93.7 &26.8 &50.2\\
ARL~\citep{shao2019facial} &59.2 &48.2 &54.9 &70.0 &83.4 &80.3 &72.0 &0.1 &0.1 &17.3 &93.0 &37.5 &51.3\\
JÂA-Net~\citep{shao2021jaa} &61.7 &50.1 &56.0 &71.7 &81.7 &82.3 &78.0 &\textbf{31.1} &1.4 &8.6 &\textbf{94.8} &37.5 &54.6\\
\cite{zhang2021prior} &\textbf{65.7} &\textbf{64.2} &\textbf{66.5} &\textbf{76.6} &74.7 &72.7 &78.6 &18.5 &10.6 &\textbf{55.1} &80.7 &\textbf{41.7} &58.8\\
AAR~\citep{shao2023facial} &65.4 &57.9 &59.9 &73.2 &\textbf{84.6} &\textbf{83.2} &\textbf{79.9} &21.8 &\textbf{27.4} &19.9 &94.5 &\textbf{41.7} &\textbf{59.1}\\
\textbf{AC$^2$D} &63.8&53.1&66.0&66.6&80.7&80.1&78.0&30.3&26.5&29.2&93.3&41.4&\textbf{59.1}\\
\bottomrule
\end{tabular}
\end{table*}

\subsubsection{Evaluation on BP4D}
The F1-frame results of our method AC$^2$D and state-of-the-art methods on BP4D are shown in Table~\ref{tab:comp_f1_bp4d}. It can be seen that our AC$^2$D achieves %the best
\highlight{good} results with average F1-frame $64.6$. unlike UGN-B, HMP-PS, \cite{jacob2021facial}, \cite{chang2022knowledge}, \highlight{and AUNet} employing external training data, AC$^2$D obtains %better
\highlight{comparable} performance using only benchmark training data. Compared to the recent causal intervention based method CISNet with additional facial identity annotations, AC$^2$D shows higher average F1-frame without depending on identity, which demonstrates the effectiveness of our proposed causal intervention on sample confounder.

\subsubsection{Evaluation on DISFA}

Table~\ref{tab:comp_f1_disfa} reports the F1-frame results on the DISFA benchmark. We can observe that our AC$^2$D outperforms \highlight{most} previous works. \highlight{Although AUNet obtains better performance than AC$^2$D, it uses additional information including pre-trained models and temporal information.} Note that there is a serious data imbalance problem in DISFA, which results in performance fluctuations across AUs for many methods like AU-GCN. In contrast, AC$^2$D achieves stable %and overall best
performance. Besides, AC$^2$D outperforms the transformer based method \cite{jacob2021facial}, which can be partially attributed to our proposed adaptive constraining on self-attention \highlight{weight} distribution. \highlight{With adaptively constrained self-attention, AC$^2$D  can precisely capture AU related local features while preserving global relational modeling ability.}

\begin{table}
\centering\caption{\highlight{Floating point operations (FLOPs) and the number of parameters (\#Params.) for typical methods during the predictions of $12$ AUs.}}
% on BP4D~\citep{zhang2014bp4d}
\label{tab:flops_params}
\setlength\tabcolsep{11pt}
% \begin{tabular}{c|c|*{2}{c}}
% \toprule
% Method &Avg & FLOPs &\#Params.\\
% \midrule
% DRML~\citep{zhao2016deep} &48.3 &\textbf{0.9G} &56.9M\\
% EAC-Net~\citep{li2018eac} &55.9 &18.8G &337.5M\\
% JÂA-Net~\citep{shao2021jaa} &62.4 &8.8G &25.2M \\
% AAR~\citep{shao2023facial} &63.8&10.2G&7.2M\\
% CISNet~\citep{chen2022causal} &64.3 &4.8G &22.4M\\
% \textbf{AC$^{2}$D} &\textbf{64.6} &9.6G &\textbf{3.6M}\\
% \bottomrule
% \end{tabular}
\begin{tabular}{c|*{2}{c}}
\toprule
Method & FLOPs &\#Params.\\
\midrule
\highlight{DRML~\citep{zhao2016deep}}  &\textbf{0.9G} &56.9M\\
\highlight{EAC-Net~\citep{li2018eac}} &18.8G &337.5M\\
\highlight{JÂA-Net~\citep{shao2021jaa}}  &8.8G &25.2M \\
\highlight{AAR~\citep{shao2023facial}} &10.2G&7.2M\\
\highlight{CISNet~\citep{chen2022causal}}  &4.8G &22.4M\\
\highlight{AUNet~\citep{yang2023toward}} &3.8G* &\textbf{2.7M}*\\
\highlight{\textbf{AC$^{2}$D}}  &9.6G &3.6M\\
\bottomrule
\end{tabular}

\begin{tablenotes}
\item \highlight{*AUNet has additional 12.1M parameters with 14.0G FLOPs in frozen pre-trained network modules.}
\end{tablenotes}
\end{table}

\begin{table*}
\centering\caption{F1-frame results for $12$ AUs of different variants of AC$^{2}$D on BP4D~\citep{zhang2014bp4d}.}
\label{tab:ablation_bp4d}
\setlength\tabcolsep{8pt}
\begin{tabular}{c|*{12}{c}|c}
\toprule
AU &1 &2 &4 &6 &7 &10 &12 &14 &15 &17 &23 &24 &\textbf{Avg}\\
\midrule
B-Net &49.8&45.9&50.3&75.0&72.7&81.6&85.5&59.8&49.0&58.3&46.3&48.5&60.2\\
$A^{v}$-Net &45.2&46.1&52.4&78.1&74.2&81.8&88.8&\textbf{63.6}&50.7&64.3&46.5&54.4&62.2\\
$A^{e}$-Net &44.0&46.8&54.3&77.9&72.9&83.9&86.0&62.6&51.5&\textbf{64.8}&48.4&49.9&61.9\\
\textbf{AC$^{2}$D} &\textbf{54.2}&\textbf{54.7}&\textbf{56.5}&77.0&76.2&\textbf{84.0}&\textbf{89.0}&\textbf{63.6}&\textbf{54.8}&63.6&46.5&\textbf{54.8}&\textbf{64.6}\\
$A^{v}C^{e(s)}$-Net &51.0 &50.2&54.6&77.7&\textbf{77.2}&82.7&88.1&60.7&52.3&64.4&\textbf{49.2}&52.3&63.4\\
$A^{v}C^{s(d)}$-Net &40.9&37.5&50.3&\textbf{78.5}&73.5&82.5&87.7&62.5&48.7&64.0&43.7&49.8&60.0\\

\bottomrule
\end{tabular}
\end{table*}

\subsubsection{Evaluation on GFT}

We present the F1-frame results on GFT in Table~\ref{tab:comp_f1_gft}. It can be observed that AC$^2$D outperforms other approaches with a large margin and improves the average F1-frame to the level $60$. Unlike BP4D and DISFA whose facial images are near-frontal, GFT images exhibit moderate out-of-plane poses. In this challenging scenario, AC$^2$D still works well.

\highlight{\subsubsection{Evaluation on BP4D+}
To evaluate the performance for testing data with larger scale and diversity, we train our AC$^2$D on the entire BP4D, and cross-dataset test on the entire BP4D+. The results of different methods are shown in Table~\ref{tab:bp4d_plus_AUocc}. It can be seen that AC$^2$D outperforms previous works in terms of average F1-frame. This demonstrates that AC$^2$D has robust performance when the scale and diversity of testing data are significantly increased.}

\subsubsection{Evaluation on Aff-Wild2}

We also compare with other methods on the challenging Aff-Wild2 benchmark in unconstrained scenarios, as presented in Table~\ref{tab:comp_f1_affwild2}. We can see that AC$^2$D achieves better performance than most of the previous works. Compared to EAC-Net and \cite{zhang2021prior} using outside training data, AC$^2$D only adopts Aff-Wild2 dataset and obtains better results. Although AC$^2$D shows comparable performance to AAR, we can notice that the results of AC$^2$D across AUs are more stable. This can be due to the separate deconfounding of sample confounder for each AU.

% Specifically, AC$^2$D obtains higher results on challenging AU 15 (lip corner depressor) and AU 24 (lip pressor), leading to more stable performance.

\highlight{\subsubsection{Discussion about Model Complexity}
Table~\ref{tab:flops_params} shows the floating point operations (FLOPs) and the number of parameters (\#Params.) of different methods for $12$ AUs. Note that many previous methods do not release the code or report FLOPs and \#Params., so we compare with methods with code or model complexity released. It can be observed that our AC$^2$D has limited number of parameters with moderate FLOPs.
When including the frozen pre-trained network modules, AC$^2$D requires less parameters and FLOPs compared to the recent work AUNet. %the least
Due to the design of a concise structure, our transformer based method is still efficient compared to previous CNNs or GNNs based methods like CISNet and AAR.
}

\begin{table}
\centering\caption{The structures of different variants of our AC$^2$D. $\mathbf{B}$: simplified ResTv2~\citep{zhang2022rest} backbone. $\mathbf{A^{v}}$: constraining on average self-attention \highlight{weight} distribution \highlight{$\mathbf{A}_i^{avg(j)}$} via $\mathcal{L}_{a}$. $\mathbf{A^{e}}$: constraining on each channel of self-attention \highlight{weight} distribution $\mathbf{A}_i^{(j)}$ via $\mathcal{L}_{a}$. $\mathbf{C^{e(d)}}$: sample deconfounding in each AU branch with sample prototype $\mathbf{f}_s^{(j)}$ extracted in a dynamic way, in which $\mathbf{f}_s^{(j)}$ is computed at each mini-batch during training. $\mathbf{C^{e(s)}}$: sample deconfounding in each AU branch with sample prototype extracted in a static way, in which sample prototype is computed at the end of each training epoch. $\mathbf{C^{s(d)}}$: sample deconfounding on shared rich feature with sample prototype extracted in a dynamic way.}%, in which sample prototype is computed at each mini-batch during training
\label{tab:variant_AC2D}
\setlength\tabcolsep{4.9pt}
\begin{tabular}{c|*{8}{c}}
\toprule
Method &$\mathbf{B}$ &$\mathbf{A^{v}}$ &$\mathbf{A^{e}}$ &$\mathbf{C^{e(d)}}$ &$\mathbf{C^{e(s)}}$ &$\mathbf{C^{s(d)}}$ &$\mathcal{L}_{u}$ &$\mathcal{L}_{a}$\\
\midrule
B-Net &$\surd$ & & & & & &$\surd$ &\\
$A^{v}$-Net &$\surd$ &$\surd$ & & & & &$\surd$ &$\surd$\\
$A^{e}$-Net &$\surd$ & &$\surd$ & & & &$\surd$&$\surd$\\
\textbf{AC$^{2}$D} &$\surd$ &$\surd$ & &$\surd$ & & &$\surd$&$\surd$\\
$A^{v}C^{e(s)}$-Net &$\surd$ &$\surd$ & & &$\surd$ & &$\surd$&$\surd$\\
$A^{v}C^{s(d)}$-Net &$\surd$ &$\surd$ & & & &$\surd$ &$\surd$&$\surd$\\
\bottomrule
\end{tabular}
\end{table}

\begin{table*}
\centering\caption{\highlight{F1-frame results for common $10$ AUs of cross evaluation between BP4D~\citep{zhang2014bp4d} and GFT~\citep{girard2017sayette}. BP4D $\rightarrow$ GFT denotes training on BP4D and testing on GFT.}}
\label{tab:ablation_bp4d_cross}
\setlength\tabcolsep{8.6pt}
\begin{tabular}{c|c|*{10}{c}|c}
\toprule
\multicolumn{2}{c|}{AU}&1&2&4&6&10&12&14&15&23&24&\textbf{Avg}\\
\midrule
\highlight{\multirow{2}*{BP4D $\rightarrow$ GFT}}
&\highlight{$A^{v}$-Net}&\textbf{28.2}&35.2&14.1&63.0&53.1&\textbf{69.4}&9.2&19.2&37.6&40.9&37.0\\
&\highlight{\textbf{AC$^{2}$D}}&28.0&\textbf{35.7}&\textbf{22.7}&\textbf{70.5}&\textbf{69.2}&65.2&\textbf{16.3}&\textbf{29.0}&\textbf{40.8}&\textbf{45.2}&\textbf{42.3}\\
\midrule
\highlight{\multirow{2}*{GFT $\rightarrow$ BP4D}}
&\highlight{$A^{v}$-Net}&42.7&38.4&9.3&\textbf{37.9}&\textbf{59.0}&\textbf{58.5}&1.0&\textbf{23.6}&36.2&32.1
&33.9\\
&\highlight{\textbf{AC$^{2}$D}}&\textbf{51.9}&\textbf{49.3}&\textbf{25.8}&24.6&50.1&40.9&\textbf{15.3}&20.1&\textbf{47.2}&\textbf{58.3}&\textbf{38.4}\\
\bottomrule
\end{tabular}
\end{table*}

\subsection{Ablation Study}

In this section, we investigate the usefulness of main components in our AC$^{2}$D framework.
Table~\ref{tab:ablation_bp4d} presents the F1-frame results of different variants of AC$^{2}$D on BP4D, in which the structure of each variant is shown in Table~\ref{tab:variant_AC2D}.
% The structures of different variants of AC$^{2}$D are shown in Table~\ref{tab:variant_AC2D}, in which the F1-frame results of each variant is presented in Table~\ref{tab:ablation_bp4d}.
B-Net uses the simplified ResTv2~\citep{zhang2022rest} backbone including stem, two stages, as well as each AU branch with the third stage followed by only one-dimensional linear layer and a Sigmoid function. Besides, it does not have the constraining on self-attention \highlight{weight} distribution.

\subsubsection{Adaptive Constraining on Self-Attention}

Based on B-Net, $A^{v}$-Net constrains the average self-attention \highlight{weight} distribution \highlight{$\mathbf{A}_i^{avg(j)}$} of the $(n_3-1)$-th block in the third stage by $\mathcal{L}_{a}$, and improves the average F1-frame from $60.2$ to $62.2$. This demonstrates the effectiveness of our proposed adaptive constraining on self-attention.
An alternative way of constraining self-attention is to encourage each channel of self-attention \highlight{weight} distribution $\mathbf{A}_i^{(j)}\in \mathbb{R}^{(k\times l/8\times l/8)\times (l/8\times l/8)}$ to close to \highlight{$\mathbf{A}_i^{gt(j)}\in \mathbb{R}^{l/8\times l/8}$}. In this case, $A^{e}$-Net obtains slightly worse performance compared to $A^{v}$-Net. This is because constraining each channel of self-attention \highlight{weight} distribution is too strict, which limits the space of self-attention learning guided by AU detection loss.

\begin{figure*}
\centering\includegraphics[width=0.951\linewidth]{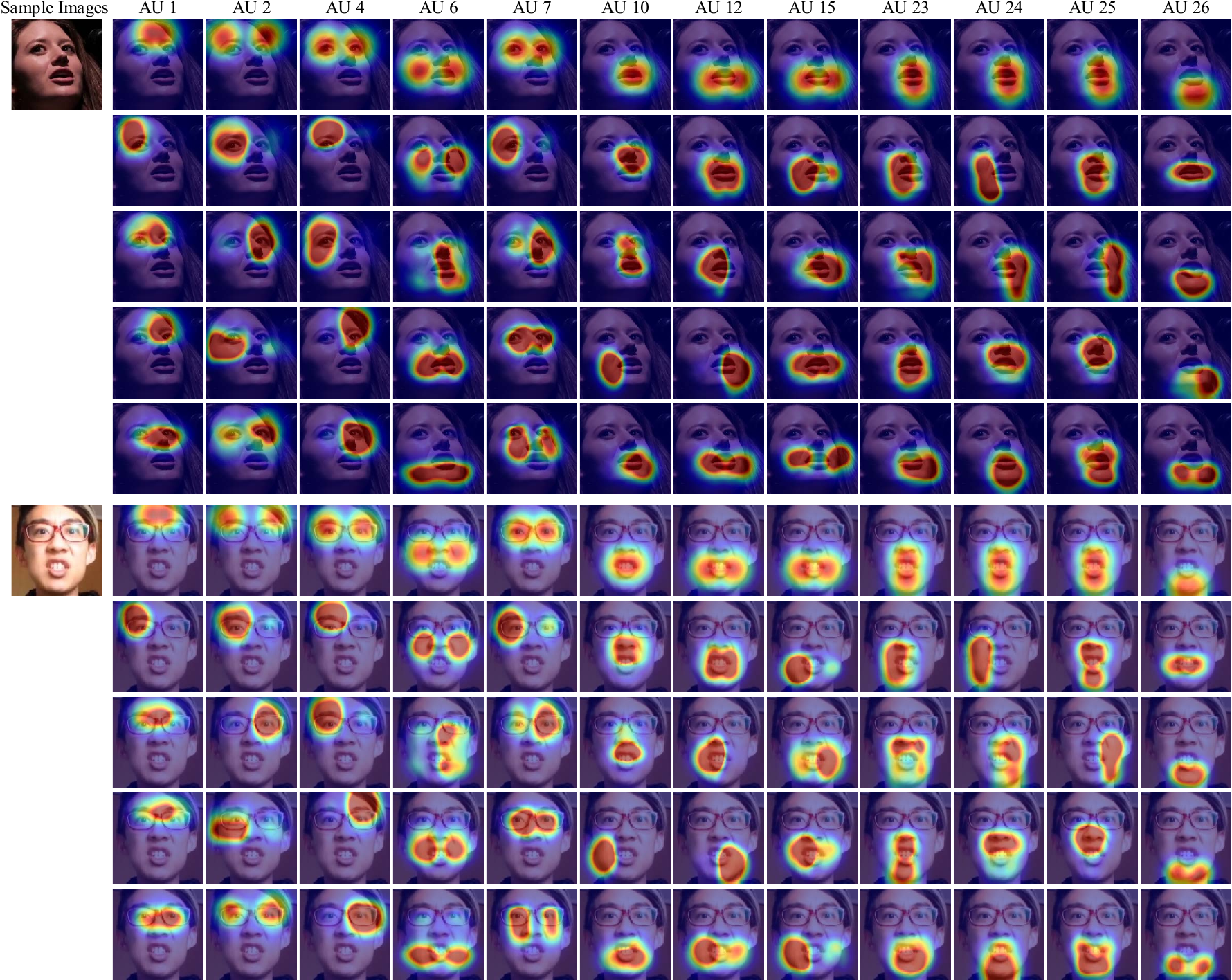}
\caption{Visualization of learned self-attention $\mathbf{A}_i^{(j)}$ by our AC$^{2}$D, in terms of the average \highlight{$\mathbf{A}_i^{avg(j)}$} and four example channels, for two sample images from Aff-Wild2~\citep{kollias2019expression,kollias2021analysing}. For each sample image, the first row shows $\mathbf{A}_i^{(j)}$ and the next four rows show randomly selected example channels. To observe the variations across samples, the two images show the same example channels. Attention weights are overlaid on the sample image for better viewing.}
\label{fig:attention_map}
\end{figure*}

\begin{figure*}
\centering\includegraphics[width=0.951\linewidth]{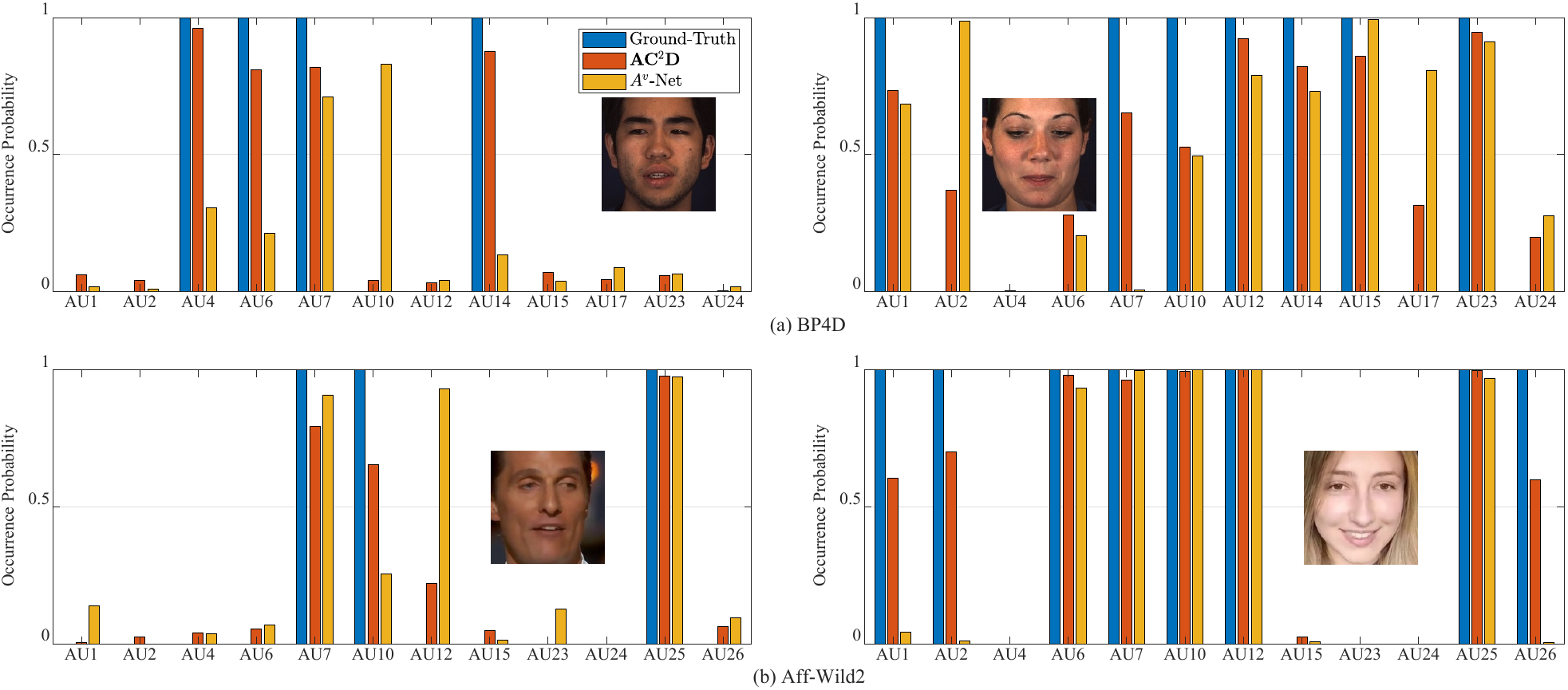}
\caption{Illustration of AU detection before and after sample deconfounding for several sample images from BP4D~\citep{zhang2014bp4d} and Aff-Wild2~\citep{kollias2019expression,kollias2021analysing}. The difference between $A^{v}$-Net and our AC$^{2}$D lies in the removal of causal intervention module.}
\label{fig:casual_prob}
\end{figure*}

\subsubsection{Causal Deconfounding of Sample Confounder}

After adding the causal intervention module in each AU branch, our AC$^{2}$D achieves the highest average F1-frame of $64.6$, in which sample prototype $z_s=\mathbf{f}_s^{(j)}$ is computed at each mini-batch during training. There are two another solutions to implement causal deconfounding of sample confounder. First, $A^{v}C^{e(s)}$-Net computes sample prototypes using current model parameters at the end of each training epoch, in which the average F1-frame is reduced to $63.4$. This is partially because generating sample prototypes in a dynamic way brings larger modelling capacity and improves robustness. Besides, computing sample prototype at each mini-batch reduces computational costs.

Second, $A^{v}C^{s(d)}$-Net adds the causal intervention module behind the shared rich feature for sample deconfounding, and obtains bad performance. There are two main reasons causing such performance degradation. A common causal intervention module for all AUs neglects AU-specific \highlight{causal} patterns. Besides, sample deconfounding on the rich feature brings large model complexity and increases the difficulty of model training.

\highlight{\subsubsection{Sample Deconfounding for Model Generalization}
To investigate the effect of sample deconfounding on model generalization ability, we compare our AC$^{2}$D with $A^{v}$-Net in terms of cross-dataset evaluation, in which the results are presented in Table~\ref{tab:ablation_bp4d_cross}. Since the evaluated $10$ AUs of GFT are all contained in the evaluated $12$ AUs of BP4D, we conduct cross-dataset evaluation between BP4D and GFT. When training on BP4D and testing on GFT, we directly use the three trained BP4D models from 3-fold cross-validation for testing and calculate the average results. Conversely, we directly test the trained GFT model on three BP4D testing sets from 3-fold cross-validation and calculate the average results.}

\highlight{
Compared to the results in Tables~\ref{tab:comp_f1_bp4d} and ~\ref{tab:comp_f1_gft}, the performance of AC$^{2}$D are significantly worse. This is due to the existing large domain gap between BP4D and GFT. Besides, we find that AC$^{2}$D works better than $A^{v}$-Net for both BP4D $\rightarrow$ GFT and GFT $\rightarrow$ BP4D. This demonstrates that our proposed causal deconfounding of sample confounder is beneficial for improving the capacity of model generalization.}

\subsection{Visual Results}

\subsubsection{Self-Attention under Adaptive Constraining}

Fig.~\ref{fig:attention_map} illustrates the visualized self-attention $\mathbf{A}_i^{(j)}$ by our AC$^{2}$D in terms of the average and a few example channels. It can be observed that the average self-attention is highlighted around the AU locations, which is beneficial for capturing AU-related region features. On the other hand, individual self-attention channels show diverse attention distributions, in which different channels model different patterns. Besides, different sample images have different distributions on the same self-attention channel, although the average self-attention \highlight{weight} distribution is similar across samples. In this case, each channel can adaptively capture potentially relevant features. %in larger spaces.
Due to the integration of both prior knowledge about AU locations and automatic self-attention learning, our proposed adaptive constraining on self-attention obtains both accurate feature learning and strong modeling ability. %are obtained.

\subsubsection{AU Detection under Sample Deconfounding}

We visualize the predicted AU occurrence probabilities \highlight{for several example images} before and after causal deconfounding of sample confounder in Fig.~\ref{fig:casual_prob}. Compared to $A^{v}$-Net without causal intervention, the predicted AU occurrence probabilities by our AC$^{2}$D are more close to the ground-truth results. \highlight{For instance,} we notice that $A^{v}$-Net predicts the co-occurrence of AU 7 (lid tightener) and AU 10 (upper lip raiser), which is not accurate for the \highlight{first} example image. Such learned AU correlation during training is often a kind of bias caused by sample characteristics. %and scenes
\highlight{Besides, $A^{v}$-Net fails to predict the co-occurrence of AU 25 (lips part) and AU 26 (jaw drop) in the fourth example image. Without sample deconfounding, it is more difficult for $A^{v}$-Net to exploit similar or relevant outside scenarios in other samples to facilitate AU detection.}
Therefore, our proposed sample deconfounding is beneficial for eliminating the \highlight{predicting} bias from sample confounder so as to improve the performance of AU detection.

\section{Conclusion}

In this paper, we have proposed a novel AU detection framework including adaptive constraining on self-attention distribution and causal deconfounding of sample confounder. In particular, we have proposed to regard the self-attention distribution of each AU as spatial distribution, and adaptively learn it under the constraint of predefined attention and the guidance of AU detection. It integrates the advantages of both prior knowledge about AU locations and automatic self-attention learning. Moreover, we have proposed to deconfound the sample confounder in the prediction of each AU by causal intervention, in which the causalities among image, sample \highlight{confounder}, and AU-specific occurrence probability are formulated. %To our knowledge, exploring the mechanism of self-attention distribution has not been done before in the AU detection field.

We have compared our approach with state-of-the-art works on the challenging BP4D, DISFA, GFT, \highlight{BP4D+}, and Aff-Wild2 benchmarks in both constrained and unconstrained scenarios. It is demonstrated that our approach %outperforms
\highlight{obtains competitive performance compared to} previous works.
%significantly in which our method obtains competitive performance. %soundly outperforms previous works.
Moreover, we have conducted an ablation study which indicates that main components in our framework all contribute to AU detection. Besides, the visual results further show the effectiveness of our proposed self-attention constraining and sample deconfounding.

\section*{Declarations}

\small{
\noindent\textbf{Author contributions}$\quad$Material preparation, data collection and analysis were mostly performed by Zhiwen Shao. The AC$^{2}$D framework was originally proposed by Zhiwen Shao, and was improved by Hancheng Zhu. Yong Zhou and Xiang Xiang, leaders of this project, delved into specific discussions of the feasibility. Bing Liu, Rui Yao, and Lizhuang Ma were involved in partial experimental designs and paper revision. The manuscript was written by Zhiwen Shao. All authors read and approved the manuscript.

\noindent\textbf{Funding}$\quad$This work was supported by the National Natural Science Foundation of China (Nos. 62472424 and 62106268), the Opening Fund of Key Laboratory of Image Processing and Intelligent Control (Huazhong University of Science and Technology), Ministry of Education, China, the Natural Science Foundation of Hubei Province (No. 2022CFB823), the China Postdoctoral Science Foundation (No. 2023M732223), and the Hong Kong Scholars Program (No. XJ2023037). It was also partially supported by the National Natural Science Foundation of China (Nos. 62101555, 62272461, 62276266, 62172417, and 72192821), the Natural Science Foundation of Jiangsu Province (Nos. BK20210488 and BK20201346), and the HUST Independent Innovation Research Fund (No. 2021XXJS096).

\noindent\textbf{Data availability}$\quad$This study uses \highlight{five} public AU datasets, including BP4D, DISFA, GFT, \highlight{BP4D+}, and Aff-Wild2. BP4D and \highlight{BP4D+} can be downloaded at http://www.cs.binghamton.edu/\url{~}lijun/Research/3DFE/3DFE\_Analysis. html, and DISFA, GFT, and Aff-Wild2 can be downloaded at http://mohammadmahoor.com/disfa, https://osf.io/7wcyz, and https://ibug.doc.ic.ac.uk/resources/aff-wild2/, respectively.
%They can be downloaded at http://www.cs.binghamton.edu/\url{~}lijun/Research/3DFE/3DFE\_Analysis. html, http://mohammadmahoor.com/disfa, https://osf.io/7wcyz, and https://ibug.doc.ic.ac.uk/resources/aff-wild2/, respectively.

\noindent\textbf{Conflict of interest}$\quad$The authors declare that the research was conducted in the absence of any commercial or financial relationships that could be construed as a potential conflict of interest.
}

% \begin{acknowledgements}
% This work was supported by the National Natural Science Foundation of China (No. 62106268), the Opening Fund of Key Laboratory of Image Processing and Intelligent Control (Huazhong University of Science and Technology), Ministry of Education, China, the China Postdoctoral Science Foundation (No. 2023M732223), the Hong Kong Scholars Program (No. XJ2023037), and the Talent Program for Deputy General Manager of Science and Technology of Jiangsu Province (No. FZ20220440). It was also partially supported by the National Natural Science Foundation of China (No. 62101555, No. 62272461, No. 62276266, No. 62172417, and No. 72192821), and the Natural Science Foundation of Jiangsu Province (No. BK20210488 and No. BK20201346).
% \end{acknowledgements}

% Authors must disclose all relationships or interests that
% could have direct or potential influence or impart bias on
% the work:
%
% \section*{Conflict of interest}
%
% The authors declare that they have no conflict of interest.
% \end{sloppypar}

% BibTeX users please use one of
\bibliographystyle{spbasic}      % basic style, author-year citations
\bibliography{references}   % name your BibTeX data base

\begin{thebibliography}{63}
\providecommand{\natexlab}[1]{#1}
\providecommand{\url}[1]{{#1}}
\providecommand{\urlprefix}{URL }
\expandafter\ifx\csname urlstyle\endcsname\relax
  \providecommand{\doi}[1]{DOI~\discretionary{}{}{}#1}\else
  \providecommand{\doi}{DOI~\discretionary{}{}{}\begingroup
  \urlstyle{rm}\Url}\fi
\providecommand{\eprint}[2][]{\url{#2}}

\bibitem[{Besserve et~al.(2020)Besserve, Mehrjou, Sun, and
  Sch{\"o}lkopf}]{besserve2020counterfactuals}
Besserve M, Mehrjou A, Sun R, Sch{\"o}lkopf B (2020) Counterfactuals uncover
  the modular structure of deep generative models. In: International Conference
  on Learning Representations

\bibitem[{Chang and Wang(2022)}]{chang2022knowledge}
Chang Y, Wang S (2022) Knowledge-driven self-supervised representation learning
  for facial action unit recognition. In: IEEE Conference on Computer Vision
  and Pattern Recognition, IEEE, pp 20417--20426

\bibitem[{Chen et~al.(2022)Chen, Chen, Wang, Wang, and Liang}]{chen2022causal}
Chen Y, Chen D, Wang T, Wang Y, Liang Y (2022) Causal intervention for
  subject-deconfounded facial action unit recognition. In: AAAI Conference on
  Artificial Intelligence, pp 374--382

\bibitem[{Chu et~al.(2017)Chu, De~la Torre, and Cohn}]{chu2017learning}
Chu WS, De~la Torre F, Cohn JF (2017) Learning spatial and temporal cues for
  multi-label facial action unit detection. In: IEEE International Conference
  on Automatic Face \& Gesture Recognition, IEEE, pp 25--32

\bibitem[{Corneanu et~al.(2018)Corneanu, Madadi, and
  Escalera}]{corneanu2018deep}
Corneanu CA, Madadi M, Escalera S (2018) Deep structure inference network for
  facial action unit recognition. In: European Conference on Computer Vision,
  Springer, pp 309--324

\bibitem[{Dosovitskiy et~al.(2021)Dosovitskiy, Beyer, Kolesnikov, Weissenborn,
  Zhai, Unterthiner, Dehghani, Minderer, Heigold, Gelly, Uszkoreit, and
  Houlsby}]{dosovitskiy2021image}
Dosovitskiy A, Beyer L, Kolesnikov A, Weissenborn D, Zhai X, Unterthiner T,
  Dehghani M, Minderer M, Heigold G, Gelly S, Uszkoreit J, Houlsby N (2021) An
  image is worth 16x16 words: Transformers for image recognition at scale. In:
  International Conference on Learning Representations

\bibitem[{Ekman and Friesen(1978)}]{ekman1978facial}
Ekman P, Friesen WV (1978) Facial action coding system: A technique for the
  measurement of facial movement. Consulting Psychologists Press

\bibitem[{Ekman et~al.(2002)Ekman, Friesen, and Hager}]{ekman2002facial}
Ekman P, Friesen WV, Hager JC (2002) Facial action coding system. Research
  Nexus

\bibitem[{Ertugrul et~al.(2020)Ertugrul, Cohn, Jeni, Zhang, Yin, and
  Ji}]{ertugrul2020crossing}
Ertugrul IO, Cohn JF, Jeni LA, Zhang Z, Yin L, Ji Q (2020) Crossing domains for
  au coding: Perspectives, approaches, and measures. IEEE Transactions on
  Biometrics, Behavior, and Identity Science 2(2):158--171

\bibitem[{Fan et~al.(2008)Fan, Chang, Hsieh, Wang, and Lin}]{fan2008liblinear}
Fan RE, Chang KW, Hsieh CJ, Wang XR, Lin CJ (2008) Liblinear: A library for
  large linear classification. Journal of Machine Learning Research
  9(Aug):1871--1874

\bibitem[{Girard et~al.(2017)Girard, Chu, Jeni, and Cohn}]{girard2017sayette}
Girard JM, Chu WS, Jeni LA, Cohn JF (2017) Sayette group formation task (gft)
  spontaneous facial expression database. In: IEEE International Conference on
  Automatic Face \& Gesture Recognition, IEEE, pp 581--588

\bibitem[{He et~al.(2017)He, Li, Yang, Cao, Sun, and Yu}]{he2017multi}
He J, Li D, Yang B, Cao S, Sun B, Yu L (2017) Multi view facial action unit
  detection based on cnn and blstm-rnn. In: IEEE International Conference on
  Automatic Face \& Gesture Recognition, IEEE, pp 848--853

\bibitem[{He et~al.(2016)He, Zhang, Ren, and Sun}]{he2016deep}
He K, Zhang X, Ren S, Sun J (2016) Deep residual learning for image
  recognition. In: IEEE Conference on Computer Vision and Pattern Recognition,
  IEEE, pp 770--778

\bibitem[{Jacob and Stenger(2021)}]{jacob2021facial}
Jacob GM, Stenger B (2021) Facial action unit detection with transformers. In:
  IEEE Conference on Computer Vision and Pattern Recognition, IEEE, pp
  7680--7689

\bibitem[{Kingma and Welling(2014)}]{diederik2014auto}
Kingma DP, Welling M (2014) Auto-encoding variational bayes. In: Bengio Y,
  LeCun Y (eds) International Conference on Learning Representations

\bibitem[{Kollias and Zafeiriou(2019)}]{kollias2019expression}
Kollias D, Zafeiriou S (2019) Expression, affect, action unit recognition:
  Aff-wild2, multi-task learning and arcface. In: British Machine Vision
  Conference, BMVA Press, p 297

\bibitem[{Kollias and Zafeiriou(2021)}]{kollias2021analysing}
Kollias D, Zafeiriou S (2021) Analysing affective behavior in the second abaw2
  competition. In: IEEE International Conference on Computer Vision Workshops,
  IEEE, pp 3652--3660

\bibitem[{Krizhevsky et~al.(2012)Krizhevsky, Sutskever, and
  Hinton}]{krizhevsky2012imagenet}
Krizhevsky A, Sutskever I, Hinton GE (2012) Imagenet classification with deep
  convolutional neural networks. In: Advances in Neural Information Processing
  Systems, Curran Associates, Inc., pp 1097--1105

\bibitem[{Kullback and Leibler(1951)}]{kullback1951information}
Kullback S, Leibler RA (1951) On information and sufficiency. The annals of
  mathematical statistics 22(1):79--86

\bibitem[{Li et~al.(2019{\natexlab{a}})Li, Zhu, Zeng, Wang, and
  Lin}]{li2019semantic}
Li G, Zhu X, Zeng Y, Wang Q, Lin L (2019{\natexlab{a}}) Semantic relationships
  guided representation learning for facial action unit recognition. In: AAAI
  Conference on Artificial Intelligence, pp 8594--8601

\bibitem[{Li et~al.(2018)Li, Abtahi, Zhu, and Yin}]{li2018eac}
Li W, Abtahi F, Zhu Z, Yin L (2018) Eac-net: Deep nets with enhancing and
  cropping for facial action unit detection. IEEE Transactions on Pattern
  Analysis and Machine Intelligence 40(11):2583--2596

\bibitem[{Li et~al.(2013)Li, Wang, Zhao, and Ji}]{li2013simultaneous}
Li Y, Wang S, Zhao Y, Ji Q (2013) Simultaneous facial feature tracking and
  facial expression recognition. IEEE Transactions on Image Processing
  22(7):2559--2573

\bibitem[{Li et~al.(2019{\natexlab{b}})Li, Zeng, Shan, and
  Chen}]{li2019self-supervised}
Li Y, Zeng J, Shan S, Chen X (2019{\natexlab{b}}) Self-supervised
  representation learning from videos for facial action unit detection. In:
  IEEE Conference on Computer Vision and Pattern Recognition, IEEE, pp
  10924--10933

\bibitem[{Liu et~al.(2022)Liu, Wang, Yang, Zhou, Yao, Shao, and
  Zhao}]{liu2022show}
Liu B, Wang D, Yang X, Zhou Y, Yao R, Shao Z, Zhao J (2022) Show, deconfound
  and tell: Image captioning with causal inference. In: IEEE Conference on
  Computer Vision and Pattern Recognition, IEEE, pp 18041--18050

\bibitem[{Liu et~al.(2020)Liu, Dong, Zhang, Wang, and Dang}]{liu2020relation}
Liu Z, Dong J, Zhang C, Wang L, Dang J (2020) Relation modeling with graph
  convolutional networks for facial action unit detection. In: International
  Conference on Multimedia Modeling, Springer, pp 489--501

\bibitem[{Lopez-Paz et~al.(2017)Lopez-Paz, Nishihara, Chintala, Scholkopf, and
  Bottou}]{lopez2017discovering}
Lopez-Paz D, Nishihara R, Chintala S, Scholkopf B, Bottou L (2017) Discovering
  causal signals in images. In: IEEE Conference on Computer Vision and Pattern
  Recognition, IEEE, pp 6979--6987

\bibitem[{Loshchilov and Hutter(2019)}]{loshchilov2017decoupled}
Loshchilov I, Hutter F (2019) Decoupled weight decay regularization. In:
  International Conference on Learning Representations

\bibitem[{Luo et~al.(2020)Luo, Shen, Cheng, Wang, and Pantic}]{luo2020shape}
Luo B, Shen J, Cheng S, Wang Y, Pantic M (2020) Shape constrained network for
  eye segmentation in the wild. In: IEEE Winter Conference on Applications of
  Computer Vision, IEEE, pp 1952--1960

\bibitem[{Ma et~al.(2019)Ma, Chen, and Yong}]{ma2019r}
Ma C, Chen L, Yong J (2019) Au r-cnn: Encoding expert prior knowledge into
  r-cnn for action unit detection. Neurocomputing 355:35--47

\bibitem[{Mavadati et~al.(2013)Mavadati, Mahoor, Bartlett, Trinh, and
  Cohn}]{mavadati2013disfa}
Mavadati SM, Mahoor MH, Bartlett K, Trinh P, Cohn JF (2013) Disfa: A
  spontaneous facial action intensity database. IEEE Transactions on Affective
  Computing 4(2):151--160

\bibitem[{Newell et~al.(2016)Newell, Yang, and Deng}]{newell2016stacked}
Newell A, Yang K, Deng J (2016) Stacked hourglass networks for human pose
  estimation. In: European Conference on Computer Vision, Springer, pp 483--499

\bibitem[{Niu et~al.(2019)Niu, Han, Yang, Huang, and Shan}]{niu2019local}
Niu X, Han H, Yang S, Huang Y, Shan S (2019) Local relationship learning with
  person-specific shape regularization for facial action unit detection. In:
  IEEE Conference on Computer Vision and Pattern Recognition, pp 11917--11926

\bibitem[{Paszke et~al.(2019)Paszke, Gross, Massa, Lerer, Bradbury, Chanan,
  Killeen, Lin, Gimelshein, Antiga, Desmaison, Kopf, Yang, DeVito, Raison,
  Tejani, Chilamkurthy, Steiner, Fang, Bai, and Chintala}]{paszke2019pytorch}
Paszke A, Gross S, Massa F, Lerer A, Bradbury J, Chanan G, Killeen T, Lin Z,
  Gimelshein N, Antiga L, Desmaison A, Kopf A, Yang E, DeVito Z, Raison M,
  Tejani A, Chilamkurthy S, Steiner B, Fang L, Bai J, Chintala S (2019)
  Pytorch: An imperative style, high-performance deep learning library. In:
  Advances in Neural Information Processing Systems, Curran Associates, Inc.,
  pp 8024--8035

\bibitem[{Pearl et~al.(2016)Pearl, Glymour, and Jewell}]{pearl2016causal}
Pearl J, Glymour M, Jewell NP (2016) Causal inference in statistics: A primer.
  John Wiley \& Sons

\bibitem[{Pearl et~al.(2000)}]{pearl2000models}
Pearl J, et~al. (2000) Models, reasoning and inference. Cambridge, UK:
  CambridgeUniversityPress 19(2):3

\bibitem[{Qi et~al.(2020)Qi, Niu, Huang, and Zhang}]{qi2020two}
Qi J, Niu Y, Huang J, Zhang H (2020) Two causal principles for improving visual
  dialog. In: IEEE Conference on Computer Vision and Pattern Recognition, IEEE,
  pp 10860--10869

\bibitem[{Rubin(2005)}]{rubin2005causal}
Rubin DB (2005) Causal inference using potential outcomes: Design, modeling,
  decisions. Journal of the American Statistical Association 100(469):322--331

\bibitem[{Sankaran et~al.(2019)Sankaran, Mohan, Setlur, Govindaraju, and
  Fedorishin}]{sankaran2019representation}
Sankaran N, Mohan DD, Setlur S, Govindaraju V, Fedorishin D (2019)
  Representation learning through cross-modality supervision. In: IEEE
  International Conference on Automatic Face \& Gesture Recognition, IEEE, pp
  1--8

\bibitem[{Shao et~al.(2021{\natexlab{a}})Shao, Liu, Cai, and Ma}]{shao2021jaa}
Shao Z, Liu Z, Cai J, Ma L (2021{\natexlab{a}}) J{\^a}a-net: Joint facial
  action unit detection and face alignment via adaptive attention.
  International Journal of Computer Vision 129(2):321--340

\bibitem[{Shao et~al.(2021{\natexlab{b}})Shao, Zhu, Tang, Lu, and
  Ma}]{shao2021explicit}
Shao Z, Zhu H, Tang J, Lu X, Ma L (2021{\natexlab{b}}) Explicit facial
  expression transfer via fine-grained representations. IEEE Transactions on
  Image Processing 30:4610--4621

\bibitem[{Shao et~al.(2022)Shao, Liu, Cai, Wu, and Ma}]{shao2019facial}
Shao Z, Liu Z, Cai J, Wu Y, Ma L (2022) Facial action unit detection using
  attention and relation learning. IEEE Transactions on Affective Computing
  13(3):1274--1289

\bibitem[{Shao et~al.(2023)Shao, Zhou, Cai, Zhu, and Yao}]{shao2023facial}
Shao Z, Zhou Y, Cai J, Zhu H, Yao R (2023) Facial action unit detection via
  adaptive attention and relation. IEEE Transactions on Image Processing
  32:3354--3366

\bibitem[{Simonyan and Zisserman(2015)}]{simonyan2014very}
Simonyan K, Zisserman A (2015) Very deep convolutional networks for large-scale
  image recognition. In: International Conference on Learning Representations

\bibitem[{Song et~al.(2021{\natexlab{a}})Song, Chen, Zheng, and
  Ji}]{song2021uncertain}
Song T, Chen L, Zheng W, Ji Q (2021{\natexlab{a}}) Uncertain graph neural
  networks for facial action unit detection. In: AAAI Conference on Artificial
  Intelligence, pp 5993--6001

\bibitem[{Song et~al.(2021{\natexlab{b}})Song, Cui, Zheng, and
  Ji}]{song2021hybrid}
Song T, Cui Z, Zheng W, Ji Q (2021{\natexlab{b}}) Hybrid message passing with
  performance-driven structures for facial action unit detection. In: IEEE
  Conference on Computer Vision and Pattern Recognition, IEEE, pp 6267--6276

\bibitem[{Szegedy et~al.(2016)Szegedy, Vanhoucke, Ioffe, Shlens, and
  Wojna}]{szegedy2016rethinking}
Szegedy C, Vanhoucke V, Ioffe S, Shlens J, Wojna Z (2016) Rethinking the
  inception architecture for computer vision. In: IEEE Conference on Computer
  Vision and Pattern Recognition, IEEE, pp 2818--2826

\bibitem[{Toisoul et~al.(2021)Toisoul, Kossaifi, Bulat, Tzimiropoulos, and
  Pantic}]{toisoul2021estimation}
Toisoul A, Kossaifi J, Bulat A, Tzimiropoulos G, Pantic M (2021) Estimation of
  continuous valence and arousal levels from faces in naturalistic conditions.
  Nature Machine Intelligence 3(1):42--50

\bibitem[{Valstar and Pantic(2006)}]{valstar2006fully}
Valstar M, Pantic M (2006) Fully automatic facial action unit detection and
  temporal analysis. In: IEEE Conference on Computer Vision and Pattern
  Recognition Workshop, IEEE, pp 149--149

\bibitem[{Vaswani et~al.(2017)Vaswani, Shazeer, Parmar, Uszkoreit, Jones,
  Gomez, Kaiser, and Polosukhin}]{vaswani_attention_2017}
Vaswani A, Shazeer N, Parmar N, Uszkoreit J, Jones L, Gomez AN, Kaiser L,
  Polosukhin I (2017) Attention is all you need. In: Advances in Neural
  Information Processing Systems, Curran Associates, Inc., pp 5998--6008

\bibitem[{Wang et~al.(2022)Wang, Qi, Cheng, and Suzuki}]{wang2022action}
Wang L, Qi J, Cheng J, Suzuki K (2022) Action unit detection by exploiting
  spatial-temporal and label-wise attention with transformer. In: IEEE
  Conference on Computer Vision and Pattern Recognition Workshops, IEEE, pp
  2470--2475

\bibitem[{Wang et~al.(2020)Wang, Huang, Zhang, and Sun}]{wang2020visual}
Wang T, Huang J, Zhang H, Sun Q (2020) Visual commonsense r-cnn. In: IEEE
  Conference on Computer Vision and Pattern Recognition, IEEE, pp 10760--10770

\bibitem[{Xu et~al.(2015)Xu, Ba, Kiros, Cho, Courville, Salakhudinov, Zemel,
  and Bengio}]{xu2015show}
Xu K, Ba J, Kiros R, Cho K, Courville A, Salakhudinov R, Zemel R, Bengio Y
  (2015) Show, attend and tell: Neural image caption generation with visual
  attention. In: International Conference on Machine Learning, PMLR, pp
  2048--2057

\bibitem[{Yang et~al.(2023)Yang, Hristov, Shen, Lin, and
  Pantic}]{yang2023toward}
Yang J, Hristov Y, Shen J, Lin Y, Pantic M (2023) Toward robust facial action
  units’ detection. Proceedings of the IEEE 111(10):1198--1214

\bibitem[{Yue et~al.(2020)Yue, Zhang, Sun, and Hua}]{yue2020interventional}
Yue Z, Zhang H, Sun Q, Hua XS (2020) Interventional few-shot learning. In:
  Advances in Neural Information Processing Systems, Curran Associates, Inc.,
  pp 2734--2746

\bibitem[{Zhang et~al.(2020)Zhang, Zhang, Tang, Hua, and Sun}]{zhang2020causal}
Zhang D, Zhang H, Tang J, Hua XS, Sun Q (2020) Causal intervention for
  weakly-supervised semantic segmentation. Advances in Neural Information
  Processing Systems 33:655--666

\bibitem[{Zhang et~al.(2019)Zhang, He, Sra, and Jadbabaie}]{zhang2019gradient}
Zhang J, He T, Sra S, Jadbabaie A (2019) Why gradient clipping accelerates
  training: A theoretical justification for adaptivity. In: International
  Conference on Learning Representations

\bibitem[{Zhang and Yang(2021)}]{zhang2021rest}
Zhang Q, Yang YB (2021) Rest: An efficient transformer for visual recognition.
  In: Advances in Neural Information Processing Systems, Curran Associates,
  Inc., pp 15475--15485

\bibitem[{Zhang and Yang(2022)}]{zhang2022rest}
Zhang Q, Yang YB (2022) Rest v2: simpler, faster and stronger. In: Advances in
  Neural Information Processing Systems, Curran Associates, Inc., pp
  36440--36452

\bibitem[{Zhang et~al.(2021)Zhang, Guo, Chen, Li, Zhang, Ding, Wu, Lv, and
  Fan}]{zhang2021prior}
Zhang W, Guo Z, Chen K, Li L, Zhang Z, Ding Y, Wu R, Lv T, Fan C (2021) Prior
  aided streaming network for multi-task affective analysis. In: IEEE
  International Conference on Computer Vision Workshops, IEEE, pp 3539--3549

\bibitem[{Zhang et~al.(2014)Zhang, Yin, Cohn, Canavan, Reale, Horowitz, Liu,
  and Girard}]{zhang2014bp4d}
Zhang X, Yin L, Cohn JF, Canavan S, Reale M, Horowitz A, Liu P, Girard JM
  (2014) Bp4d-spontaneous: A high-resolution spontaneous 3d dynamic facial
  expression database. Image and Vision Computing 32(10):692--706

\bibitem[{Zhang et~al.(2016)Zhang, Girard, Wu, Zhang, Liu, Ciftci, Canavan,
  Reale, Horowitz, Yang, Cohn, Ji, and Yin}]{zhang2016multimodal}
Zhang Z, Girard JM, Wu Y, Zhang X, Liu P, Ciftci U, Canavan S, Reale M,
  Horowitz A, Yang H, Cohn JF, Ji Q, Yin L (2016) Multimodal spontaneous
  emotion corpus for human behavior analysis. In: IEEE Conference on Computer
  Vision and Pattern Recognition, IEEE, pp 3438--3446

\bibitem[{Zhao et~al.(2016{\natexlab{a}})Zhao, Chu, De~la Torre, Cohn, and
  Zhang}]{zhao2016joint}
Zhao K, Chu WS, De~la Torre F, Cohn JF, Zhang H (2016{\natexlab{a}}) Joint
  patch and multi-label learning for facial action unit and holistic expression
  recognition. IEEE Transactions on Image Processing 25(8):3931--3946

\bibitem[{Zhao et~al.(2016{\natexlab{b}})Zhao, Chu, and Zhang}]{zhao2016deep}
Zhao K, Chu WS, Zhang H (2016{\natexlab{b}}) Deep region and multi-label
  learning for facial action unit detection. In: IEEE Conference on Computer
  Vision and Pattern Recognition, IEEE, pp 3391--3399

\end{thebibliography}

% % Non-BibTeX users please use
% \begin{thebibliography}{}
% %
% % and use \bibitem to create references. Consult the Instructions
% % for authors for reference list style.
% %
% \bibitem{RefJ}
% % Format for Journal Reference
% Author, Article title, Journal, Volume, page numbers (year)
% % Format for books
% \bibitem{RefB}
% Author, Book title, page numbers. Publisher, place (year)
% % etc
% \end{thebibliography}

\end{sloppypar}
\end{document}